\documentclass{article}

     \usepackage[final]{neurips_2019}

\title{
Likelihood Ratios for Out-of-Distribution Detection 
}

\author{%
  Jie Ren\thanks{Corresponding authors%
  }~~\thanks{Google AI Resident}\\
  Google Research \\
  \texttt{jjren@google.com} 
  \And
  Peter J. Liu \thanks{Mentors}\\
  Google Research \\
  \texttt{peterjliu@google.com} 
  \And
  Emily Fertig\footnotemark[2] \\
  Google Research \\
  \texttt{emilyaf@google.com} 
  \And
  Jasper Snoek \\
  Google Research \\
  \texttt{jsnoek@google.com} 
  \And
  Ryan Poplin \\
  Google Research  \\
  \texttt{rpoplin@google.com} \\
  \And
  Mark A. DePristo \\
  Google Research  \\
  \texttt{mdepristo@google.com} \\
  \And
  Joshua V. Dillon \footnotemark[3] \\
  Google Research  \\
  \texttt{jvdillon@google.com} \\
  \And
  Balaji Lakshminarayanan\footnotemark[1] \ \footnotemark[3] \\
  DeepMind  \\
  \texttt{balajiln@google.com} \\
}

\usepackage{algorithm,algorithmic}      %

\usepackage[utf8]{inputenc} %
\usepackage[T1]{fontenc}    %

\usepackage{url}            %
\usepackage{booktabs}       %
\usepackage{amsfonts}       %
\usepackage{amsmath}
\usepackage{amssymb}
\usepackage{natbib}
\usepackage{nicefrac}       %
\usepackage{microtype}      %
\usepackage{graphicx}
\usepackage{subfigure}
\usepackage{adjustbox}
\usepackage{wrapfig}
\usepackage{tabularx}
\usepackage{multirow}
\usepackage{bm}
\usepackage{bbm}
\usepackage{color}
\usepackage{enumitem}
\setlist[enumerate]{leftmargin=15pt}
\setlist[itemize]{leftmargin=15pt}

\definecolor{mydarkblue}{rgb}{0,0.08,0.45}
\usepackage[colorlinks,citecolor=mydarkblue,urlcolor=mydarkblue,linkcolor=mydarkblue,filecolor=mydarkblue]{hyperref}

\usepackage{amsmath,amsfonts,bm}

\def\eqref#1{equation~\ref{#1}}
\def\Eqref#1{Equation~\ref{#1}}

\def\1{\bm{1}}

\def\vtheta{{\bm{\theta}}}

\def\vx{{\bm{x}}}

\def\vz{{\bm{z}}}

\DeclareMathAlphabet{\mathsfit}{\encodingdefault}{\sfdefault}{m}{sl}
\SetMathAlphabet{\mathsfit}{bold}{\encodingdefault}{\sfdefault}{bx}{n}

\newcommand{\E}{\mathbb{E}}

\newcommand{\Var}{\mathrm{Var}}

\newcommand{\llr}{\mathsf{LLR}}

\newcommand{\myvspace}[1]{\vspace{#1}} %

\begin{document}
 \maketitle %

\begin{abstract}
Discriminative neural networks offer little or no performance guarantees when deployed on data not generated by the same process as the training distribution. On such out-of-distribution (OOD) inputs, the prediction may not only be erroneous, but confidently so, limiting the safe deployment of classifiers in real-world applications. One such challenging application is bacteria identification based on genomic sequences, which holds the promise of early detection of diseases, but requires a model that can output low confidence predictions on OOD genomic sequences from new bacteria that were not present in the training data.  We introduce a genomics dataset for OOD detection that allows other researchers to benchmark progress on this important problem. 
We investigate deep generative model based approaches for OOD detection and observe that the likelihood score is heavily affected by population level background statistics. We propose a likelihood ratio method for deep generative models which effectively corrects for these confounding background statistics. We benchmark the OOD detection performance of the proposed method against existing approaches on the genomics dataset and show that our method achieves state-of-the-art performance. 
We demonstrate the generality of the proposed method by showing that it significantly improves OOD detection when applied to deep generative models of images. 

\end{abstract}

\section{Introduction}
\label{intro}
For many machine learning systems, being able to detect data that is anomalous or significantly different from that used in training can be critical to maintaining safe and reliable predictions.  This is particularly important for deep neural network classifiers which have been shown to %
incorrectly classify such \emph{out-of-distribution} (OOD) inputs into in-distribution classes with high confidence~\citep{goodfellow6572explaining,nguyen2015deep}. 
This behaviour can have serious consequences when the predictions inform real-world decisions such as medical diagnosis, e.g.\ falsely classifying a healthy sample as pathogenic or vice versa can have extremely high cost. The importance of dealing with OOD inputs, also referred to as distributional shift, has been recognized as an important problem for AI safety \citep{amodei2016concrete}.
The majority of recent work on OOD detection for neural networks is evaluated on image datasets where the neural network is trained on one benchmark dataset (e.g. CIFAR-10) and tested on another (e.g. SVHN). While these benchmarks are important, there is a need for more realistic datasets which reflect the challenges of dealing with OOD inputs in practical applications. 

Bacterial identification is one of the most important sub-problems of many types of medical diagnosis. 
For example, diagnosis and treatment of infectious diseases, such as sepsis, relies on the accurate detection of bacterial infections in blood \citep{blauwkamp2019analytical}.
Several machine learning methods have been developed to perform bacteria identification by classifying existing known genomic sequences \citep{patil2011taxonomic,rosen2010nbc}, including deep learning methods \citep{busia2018deep} which are state-of-the-art.
 Even if neural network classifiers achieve high accuracy  as measured through cross-validation, deploying them is challenging as real data is highly likely to contain genomes from unseen classes not present in the training data.  
{Different bacterial classes continue to be discovered gradually over the years (see Figure~\ref{fig:accumu} in Appendix~\ref{dataset}) and it is estimated that 60\%-80\% of genomic sequences belong to as yet unknown} bacteria~\citep{zhu2018metagenomic, eckburg2005diversity, nayfach2019new}.  
Training a classifier on existing bacterial classes and deploying it may result in 
OOD inputs being wrongly classified as one of the classes from the training data with high confidence. 
 In addition, OOD inputs can also be the contaminations from the bacteria's host genomes such as human, plant, fungi, etc., which also need to be detected and excluded from predictions \citep{ponsero2019promises}. %
Thus having a method for accurately detecting OOD inputs is critical to enable the practical application of machine learning methods to this important problem.

A popular and intuitive 
strategy for detecting OOD inputs is to train a generative model (or a hybrid model cf.~\cite{nalisnick2019hybrid}) on training data and use that to detect OOD inputs at test time  \citep{bishop1994novelty}. However, \citet{nalisnick2018deep} and \citet{choi2018waic} recently showed that deep generative models trained on image datasets can assign higher likelihood to OOD inputs. 
We report a similar failure mode for likelihood based OOD detection using deep generative models of genomic sequences. We investigate this phenomenon and find that the likelihood can be confounded by general population level background statistics.
We propose a likelihood ratio method which uses a background model to correct for the background %
statistics and enhances the %
in-distribution specific features for OOD detection. While our investigation was motivated by the genomics problem, we found our methodology to be more general and it shows positive results on image datasets as well. 
In summary, our contributions are:
\begin{itemize}
    \item We create a realistic benchmark for OOD detection, that is motivated by challenges faced in applying deep learning models on genomics data. The sequential nature of genetic sequences provides a new modality and hopefully encourages the OOD research community to contribute to ``machine learning that matters''~\citep{wagstaff-2012}.
    \item {We show that likelihood from deep generative models can be confounded by background statistics.}
    \item We propose a %
    likelihood ratio method for OOD detection, which significantly outperforms the raw likelihood on OOD detection for deep generative models on image datasets. %
    \item We evaluate existing OOD methods on the proposed genomics benchmark and demonstrate that our method achieves  state-of-the-art (SOTA) performance on this challenging problem.
\end{itemize}

\section{Background}

\begin{figure}[ht]
\begin{center}
     \includegraphics[width=0.325\textwidth]{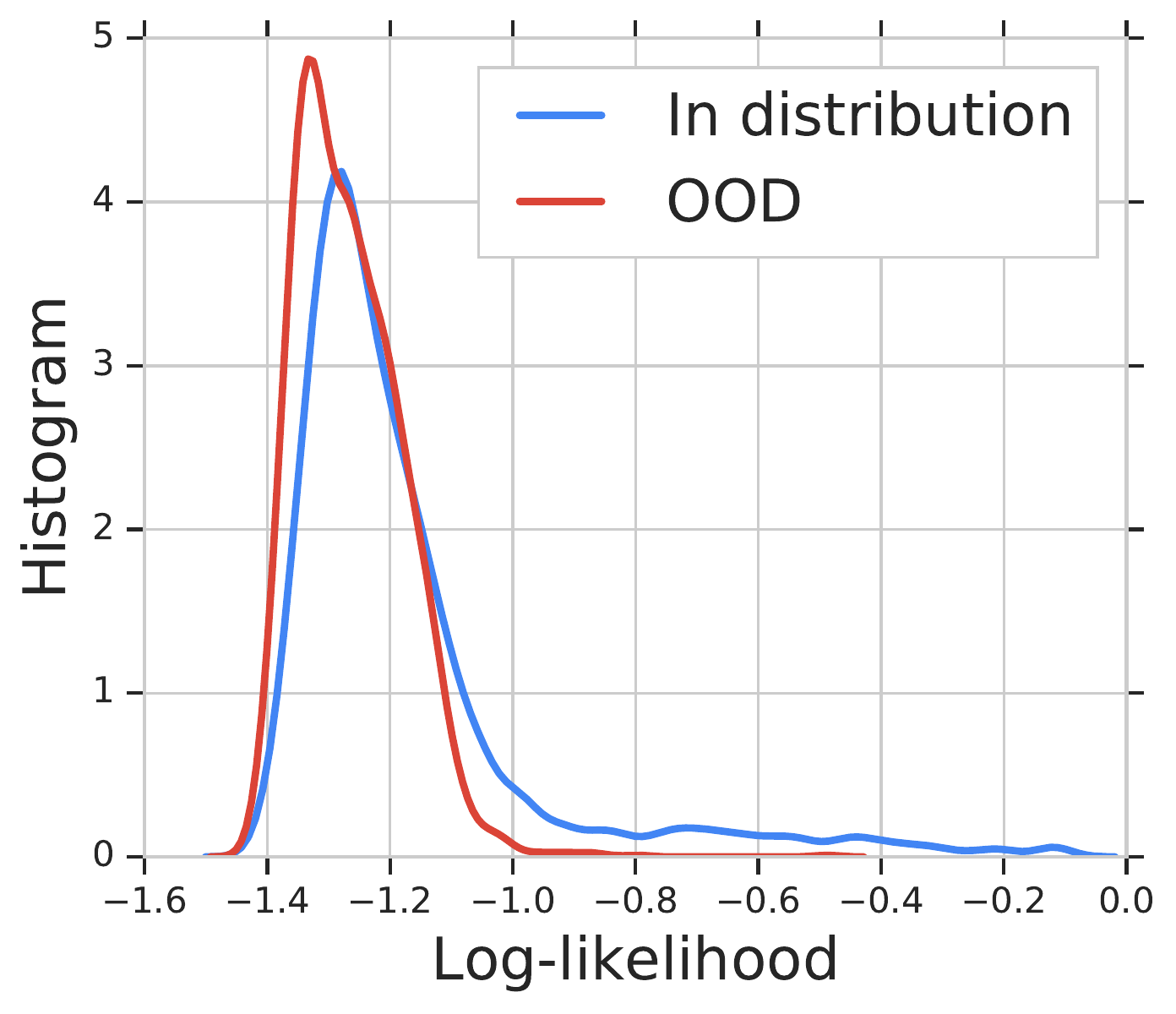}
     \includegraphics[width=0.325\textwidth]{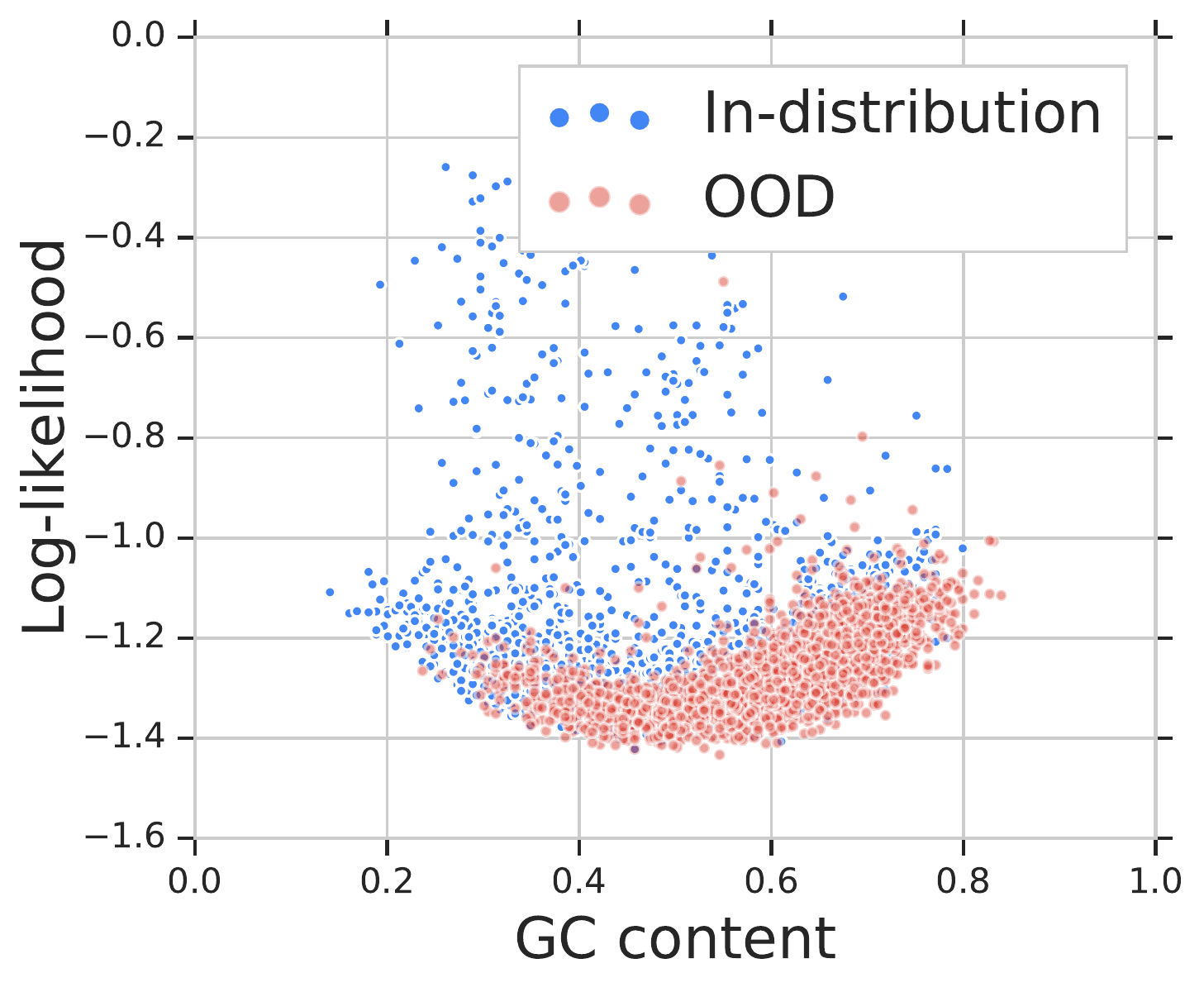}
    \caption{(a) Log-likelihood hardly separates in-distribution and OOD inputs with AUROC of 0.626. (b) The log-likelihood is heavily affected by the GC-content of a sequence. %
    }
    \label{fig:gc_before}
\end{center}    
\end{figure}

Suppose we have an in-distribution dataset $\mathcal{D}$  of $(\vx, y)$ pairs sampled from the distribution $p^*(\vx, y)$, where $\vx$ is the extracted feature vector or raw input and $y \in \mathcal{Y} := \{1, \dots, k, \dots, K\}$ is the label assigning membership to one of $K$ in-distribution classes. For simplicity, we assume inputs to be discrete, i.e.~$x_d\in\{A,C,G,T\}$ for genomic sequences and $x_d\in\{0,\dots,255\}$ for images. In general, OOD inputs are samples $(\vx, y)$ generated from an underlying distribution other than $p^*(\vx, y)$. %
In this paper, we consider an input $(\vx, y)$ to be OOD if $y \not\in \mathcal{Y}$: that is, the class $y$ does not belong to one of the $K$ in-distribution classes. Our goal is to accurately detect if an input $\vx$ is OOD or not. 
Many existing methods involve computing statistics using the predictions of (ensembles of) discriminative classifiers trained on in-distribution data, e.g.\ taking the confidence or entropy of the predictive distribution $p(y|\vx)$~\citep{hendrycks2016baseline,lakshminarayanan2017simple}. %
An alternative is to use generative model-based methods, which %
are appealing as they do not require labeled data and directly model the input distribution.  These methods fit a  generative model $p(\vx)$ to the input data, 
and then evaluate the likelihood of new inputs under that model.
However, recent work has highlighted significant issues with this approach for OOD detection on images, showing that deep generative models such as Glow~\citep{kingma2018glow} and PixelCNN~\citep{oord2016pixel,salimans2017pixelcnn++} sometimes assign higher likelihoods to OOD than in-distribution inputs.  
For example, \citet{nalisnick2018deep} and \citet{choi2018waic} %
show that Glow models trained on the CIFAR-10 image dataset assign higher likelihood to OOD inputs from the SVHN dataset than they do to in-distribution CIFAR-10 inputs;   \citet{nalisnick2018deep}, \citet{shafaei2018does} and \citet{hendrycks2018deep} show failure modes of PixelCNN and PixelCNN++ for OOD detection. %
\textbf{Failure of density estimation for OOD detection}
We investigate whether density estimation-based methods work well for OOD detection in genomics.
As a motivating observation, we train a deep generative model, more precisely  LSTM \citep{lstm}, on in-distribution genomic sequences (composed by \{A, C, G, T\}), and plot the log-likelihoods of both in-distribution and OOD inputs (See Section~\ref{sec:genomics} for the dataset and the full experimental details).
Figure~\ref{fig:gc_before}a shows that the histogram of the log-likelihood for OOD sequences largely overlaps with that of in-distribution sequences with AUROC of 0.626, making it unsuitable for OOD detection.
Our observations show a failure mode of deep generative models for OOD detection on genomic sequences 
and are complementary to earlier work which showed similar results for deep generative models on images \citep{nalisnick2018deep, choi2018waic}. 

When investigating this failure mode, %
we discovered that the log-likelihood under the model is heavily affected by a sequence's \emph{GC-content}, see Figure~\ref{fig:gc_before}b. 
GC-content is defined as the percentage of bases that are either G or C,  %
and is used widely in genomic studies as a basic statistic for describing overall genomic composition \citep{sueoka1962genetic}, and studies have shown that bacteria have an astonishing diversity of genomic GC-content, from 16.5\% to 75\% \citep{hildebrand2010evidence}.  
Bacteria from similar groups tend to have similar GC-content at the population level, but they also have characteristic biological patterns that can distinguish them well from each other.  
The confounding effect of GC-content in Figure~\ref{fig:gc_before}b makes the likelihood less reliable as a score for OOD detection, because an 
OOD input may result in a higher likelihood than an in-distribution input, because it has high GC-content (cf.~the bottom right part of Figure~\ref{fig:gc_before}b) and not necessarily because it contains characteristic  patterns specific to the in-distribution bacterial classes.

\section{Likelihood Ratio for OOD detection}
We first describe the high level idea and then describe how to adapt it to deep generative models.

\textbf{High level idea} Assume that an input $\vx$ is composed of two components, (1) a \textit{background} component characterized by population level background statistics, and (2) a \textit{semantic} component characterized by  patterns specific to the in-distribution data. 
For example, images can be modeled as backgrounds plus objects; text can be considered as a combination of high frequency stop words plus semantic words \citep{luhn1960key}; genomes can be modeled as background sequences plus motifs \citep{bailey1995value, reinert2009alignment}. 
More formally, for a $D$-dimensional input $\vx=x_1, \ldots, x_D$, we assume that there exists an unobserved variable $\vz=z_1,\ldots, z_D$, where $z_d \in \{B, S\}$ indicates if the $d$th dimension of the input $x_d$ is generated from the \textit{B}ackground model or the \textit{S}emantic model. Grouping the semantic and background parts, the input can be factored as $\vx=\{\vx_B,\vx_S\}$ where $\vx_B = \{x_d \mid z_d = B, d=1, \ldots, D \}$. 
For simplicity, assume that the background and semantic components are generated independently. 
The likelihood can be then decomposed as follows,
\begin{align}\label{eq:factorization}
p(\vx) = p(\vx_B)  p(\vx_S). 
\end{align}
When training and evaluating deep generative models, we typically do not distinguish between these two terms in the likelihood. However, we may want to use just the semantic likelihood $p(\vx_S)$ to avoid the likelihood term being dominated by the background term (e.g. OOD input with the same background but different semantic component). %
In practice, we only observe $\vx$, and it is not always easy to split an input into background and semantic parts $\{\vx_B,\vx_S\}$. %
As a practical alternative, we propose training a background model by perturbing inputs. Adding the right amount of perturbations to inputs can corrupt the semantic structure in the data, and hence the model trained on perturbed inputs captures only the population level background statistics. 

Assume that $p_\vtheta(\cdot)$ is a  model trained using in-distribution data, and $p_{\vtheta_0}(\cdot)$ is a background model that captures general background statistics. 
We propose a likelihood ratio statistic that is defined as
\begin{align}\label{eq:lr}
\llr(\vx) = \log \frac{p_\vtheta(\vx)}{p_{\vtheta_0}(\vx)} 
= \log \frac{p_\vtheta(\vx_{B}) \ p_\vtheta(\vx_{S})}{p_{\vtheta_0}(\vx_{B})\ p_{\vtheta_0}(\vx_{S})},
\end{align}
where we use the factorization from \Eqref{eq:factorization}. 
Assume that (i) both models capture the background information equally well, that is $p_{\vtheta}(\vx_{B}) \approx p_{\vtheta_0}(\vx_{B})$ and (ii)
   $p_\vtheta(\vx_{S})$ is more peaky than $p_{\vtheta_0}(\vx_{S})$ as the former is trained on data containing semantic information, while the latter model $\vtheta_0$ is trained using data with noise perturbations. 
Then, the likelihood ratio can be approximated as 
\begin{align}
\llr(\vx) \approx \log p_\vtheta(\vx_{S}) - \log p_{\vtheta_0}(\vx_{S}).
\end{align}

After taking the ratio, the likelihood for the background component $\vx_{B}$ is cancelled out, and only the likelihood %
 for the semantic component $\vx_{S}$ remains.  %
Our method produces a \textit{background contrastive} score that captures the significance of the semantics compared with the background model. 

\textbf{Likelihood ratio for auto-regressive models} 
Auto-regressive models are one of the popular choices for generating images \citep{oord2016pixel, van2016conditional,salimans2017pixelcnn++} and sequence data such as genomics \citep{zou2018primer, killoran2017generating} and drug molecules \citep{olivecrona2017molecular, gupta2018generative}, and text \citep{jozefowicz2016exploring}.
In auto-regressive models, the log-likelihood of an input can be expressed as  $\log p_{\vtheta}(\vx)=\sum_{d=1}^{D} \log p_{\vtheta} (x_d|\vx_{<d})$, 
where $\vx_{<d}=x_1\ldots x_{d-1}$. Decomposing the log-likelihood into background and semantic parts, we have
\begin{align}
    \log p_{\vtheta}(\vx)=\sum_{d: x_d \in \vx_{B}} \log p_{\vtheta} (x_d|\vx_{<d}) + \sum_{d:x_d \in \vx_{S}} \log p_{\vtheta} (x_d|\vx_{<d}).
\end{align}
We can use a similar auto-regressive decomposition for the background model $p_{\vtheta_0}(\vx)$ as well. Assuming that both the models capture the background information equally well, $\sum_{d:x_d \in \vx_{B}} \log p_{\vtheta} (x_d|\vx_{<d}) \approx \sum_{d:x_d \in \vx_{B}} \log p_{\vtheta_0} (x_d|\vx_{<d})$, the likelihood ratio is approximated as
\begin{align}
\llr(\vx) \approx \sum_{d:x_d \in \vx_{S}} \log p_{\vtheta} (x_d|\vx_{<d}) - \sum_{d:x_d \in \vx_{S}}  \log p_{\vtheta_0} (x_d|\vx_{<d}) = \sum_{d:x_d \in \vx_{S}} \log \frac{p_{\vtheta} (x_d|\vx_{<d})}{p_{\vtheta_0} (x_d|\vx_{<d})}.
\end{align}

\textbf{Training the Background Model} In practice, we add perturbations to the input data by randomly selecting positions in $x_1\ldots x_D$ following an independent and identical Bernoulli distribution with rate $\mu$ and substituting the original character with one of the other characters with equal probability. %
The procedure is inspired by genetic mutations. 
See Algorithm~\ref{alg:noise} in Appendix~\ref{sec:pseudocode} for the pseudocode for generating input perturbations. %
The rate $\mu$
is a hyperparameter and can be easily tuned using a small amount of validation OOD dataset (different from the actual OOD dataset of interest).  
In the case where validation OOD dataset is not available, we show that $\mu$ can also be tuned using simulated OOD data.
In practice, we observe that $\mu \in [0.1,0.2]$ achieves good performance empirically for most of the experiments in our paper.  
Besides adding perturbations to the input data, we found other techniques that can improve model generalization and prevent model memorization, such as adding $L_2$ regularization with coefficient $\lambda$
to model weights, can help to train a good background model. In fact, it has been shown that adding noise to the input is equivalent to adding $L_2$ regularization to the model weights under some conditions \citep{bishop1995regularization, bishop1995training}. 
Besides the methods above, we expect adding other types of noise or regularization methods would show a similar effect.
The pseudocode for our proposed OOD detection algorithm can be found in Algorithm \ref{alg:llr} in Appendix~\ref{sec:pseudocode}. 

\section{Experimental setup} 
We design  experiments on multiple data modalities (images, genomic sequences) to evaluate our method and compare with other baseline methods. 
For each of the datasets, we build an auto-regressive model for computing the log-likelihood $\log p_\vtheta(\vx)=\sum_{d=1}^{D} \log p_{\vtheta}(x_d|\vx_{<d})$.
For training the background model $p_{\vtheta_0}(\vx)$, we use the exact same architecture as  $p_{\vtheta}(\vx)$, and the only differences are that it is trained on perturbed inputs and (optionally) we apply $L_2$ regularization to model weights. 

\textbf{Baseline methods for comparison}
We compare our approach to several existing methods. %
\begin{enumerate}
    \item The maximum class probability, $p(\hat{y}|\vx)=\max_k p(y=k|\vx)$.  %
    OOD inputs tend to have lower scores than in-distribution data~\citep{hendrycks2016baseline}.
    \item The entropy of the predicted class distribution, $-\sum_{k} p(y=k|\vx) \log p(y=k|\vx)$. High entropy 
    of the predicted class distribution, and therefore a 
    high predictive uncertainty, which suggests that the input may be OOD.
    \item The ODIN method proposed by \citet{liang2017enhancing}. ODIN uses temperature scaling \citep{guo2017calibration}, adds small perturbations to the input, and applies a threshold to the resulting predicted class to distinguish in- and out-of- distribution inputs. This method was designed for continuous inputs and cannot be directly applied to discrete genomic sequences. We propose instead to add perturbations to the input of the last layer that is closest to the output of the neural network. 
    \item The Mahalanobis distance of the input to the nearest class-conditional Gaussian distribution estimated from the in-distribution data. \citet{lee2018simple} fit  class-conditional Gaussian distributions 
    to the activations from the last layer of the neural network. %
    \item The classifier-based ensemble method that uses the average of the predictions from multiple independently trained models with random initialization of network parameters and random shuffling of training inputs \citep{lakshminarayanan2017simple}. 
    \item The log-odds of a binary classifier trained to distinguish between in-distribution inputs from all classes as one class and randomly perturbed in-distribution inputs as the other. 
    \item The maximum class probability over $K$ in-distribution classes of a $(K+1)$-class classifier where the additional class is perturbed in-distribution.
    \item The maximum class probability of a $K$-class classifier for in-distribution classes but the predicted class distribution is explicitly trained to output %
    uniform distribution on perturbed in-distribution inputs. This is similar to  using simulated OOD inputs from GAN \citep{lee2017training} or using auxiliary datasets of outliers \citep{hendrycks2018deep} %
    for calibration purpose. %
    \item The generative model-based ensemble method that measures 
    $\E[\log p_\vtheta(\vx)]-\Var[\log p_\vtheta(\vx)]$ 
    of multiple likelihood estimations from independently trained model with random initialization and random shuffling of the inputs. \citep{choi2018waic}. 
\end{enumerate}

Baseline methods 1-8 are classifier-based and method 9 is generative model-based.
For classifier-based methods, we choose the commonly used model architecture, convolutional neural networks (CNNs). %
Methods 6-8 are based on perturbed inputs which aims to mimic OOD inputs. Perturbations are added to the input in the same way as that we use for training background models. 
Our method and methods 3, 6, 7, and 8 involve hyperparameter tuning; %
we follow the protocol of \citet{hendrycks2018deep} where optimal hyperparameters are picked on a different OOD validation set than the final OOD dataset it is tested on. %
For Fashion-MNIST vs. MNIST experiment, we use the  NotMNIST \cite{notmnist} dataset for hyperparameter tuning. For CIFAR-10 vs SVHN, we used gray-scaled CIFAR-10 for hyperparameter tuning. For genomics, we use the OOD bacteria classes discovered between 2011-2016, which are disjoint from the final OOD classes discovered after 2016. 
While this set of baselines is not exhaustive, it is broadly representative of the range of existing methods. 
{Note that since our method does not rely on OOD inputs for training, we do not compare it with other %
methods that do utilize OOD inputs in training.}

\textbf{Evaluation metrics for OOD detection}
We trained the model using only in-distribution inputs, and we tuned the hyperparameters using validation datasets that include both in-distribution and OOD inputs. The test dataset is used for the final evaluation of the method. 
For the final evaluation, we randomly selected the same number of in-distribution and OOD inputs from the test dataset, and for each example $\vx$ we computed the log likelihood-ratio statistic $\llr(\vx)$ as the score. 
A small value of the score suggests a high likelihood of being OOD. We use the area under the ROC curve (AUROC$\uparrow$), the area under the precision-recall curve (AUPRC$\uparrow$), and the false positive rate at 80\% true positive rate (FPR80$\downarrow$), as the metrics for evaluation. These three metrics are commonly used for evaluating OOD detection methods \citep{hendrycks2016baseline, hendrycks2018deep, alemi2018uncertainty}. 

\section{Results}
We first present results on image datasets as they are easier to visualize, and then present results on our proposed genomic dataset. For image experiments, our goal is not to achieve  state-of-the-art performance but to show that our likelihood ratio effectively corrects for background statistics and significantly outperforms the likelihood. While previous work has shown the failure of PixelCNN for OOD detection, we believe ours is the first to provide an explanation for why this phenomenon happens for PixelCNN, through the lens of background statistics.  

\subsection{Likelihood ratio for detecting OOD images}

\begin{figure}[h]%
    \centering
    \myvspace{-1em}
    \begin{subfigure}[]{
    \includegraphics[width=0.325\textwidth]{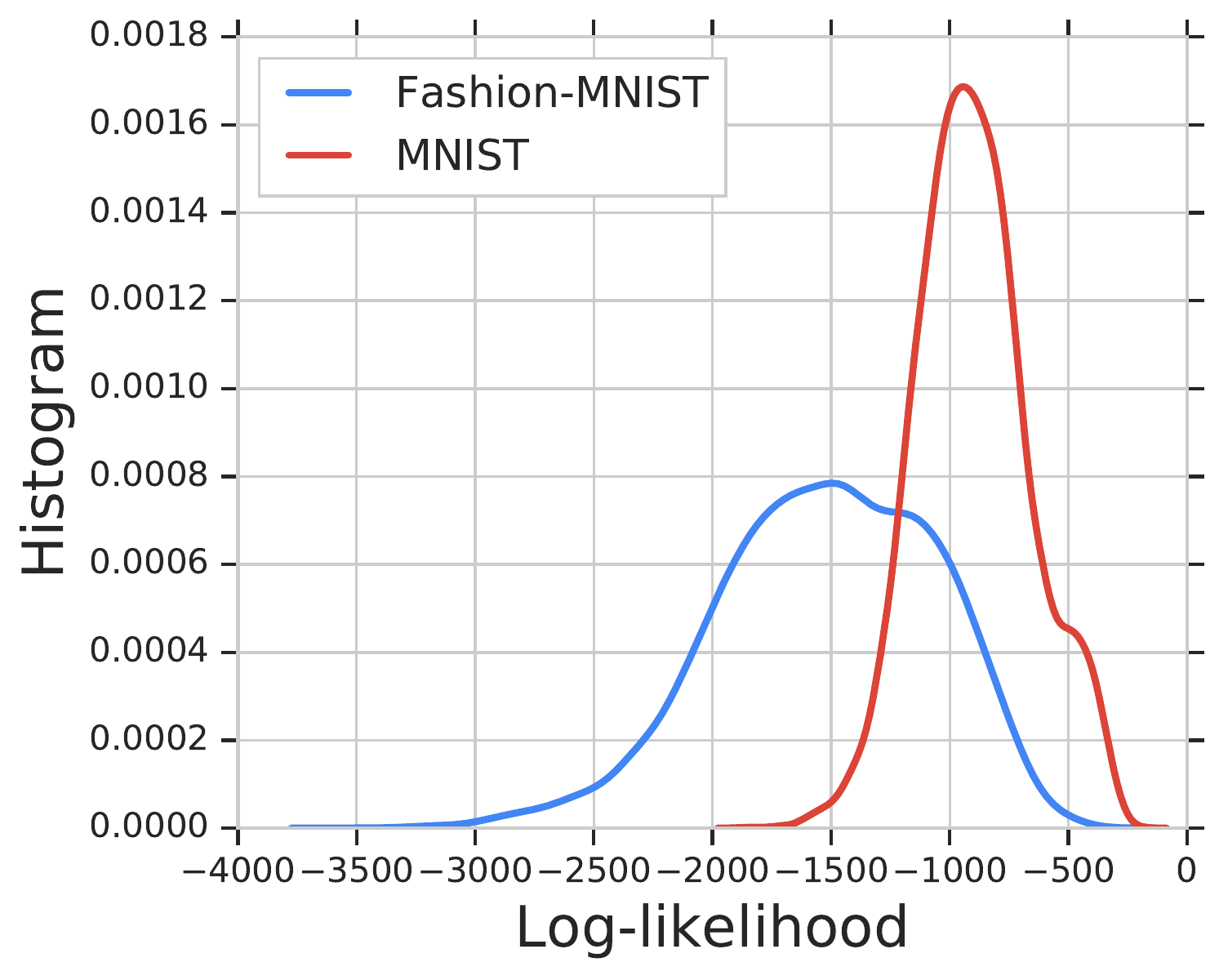}
    }\end{subfigure}
    \begin{subfigure}[]{
    \includegraphics[width=0.325\textwidth]{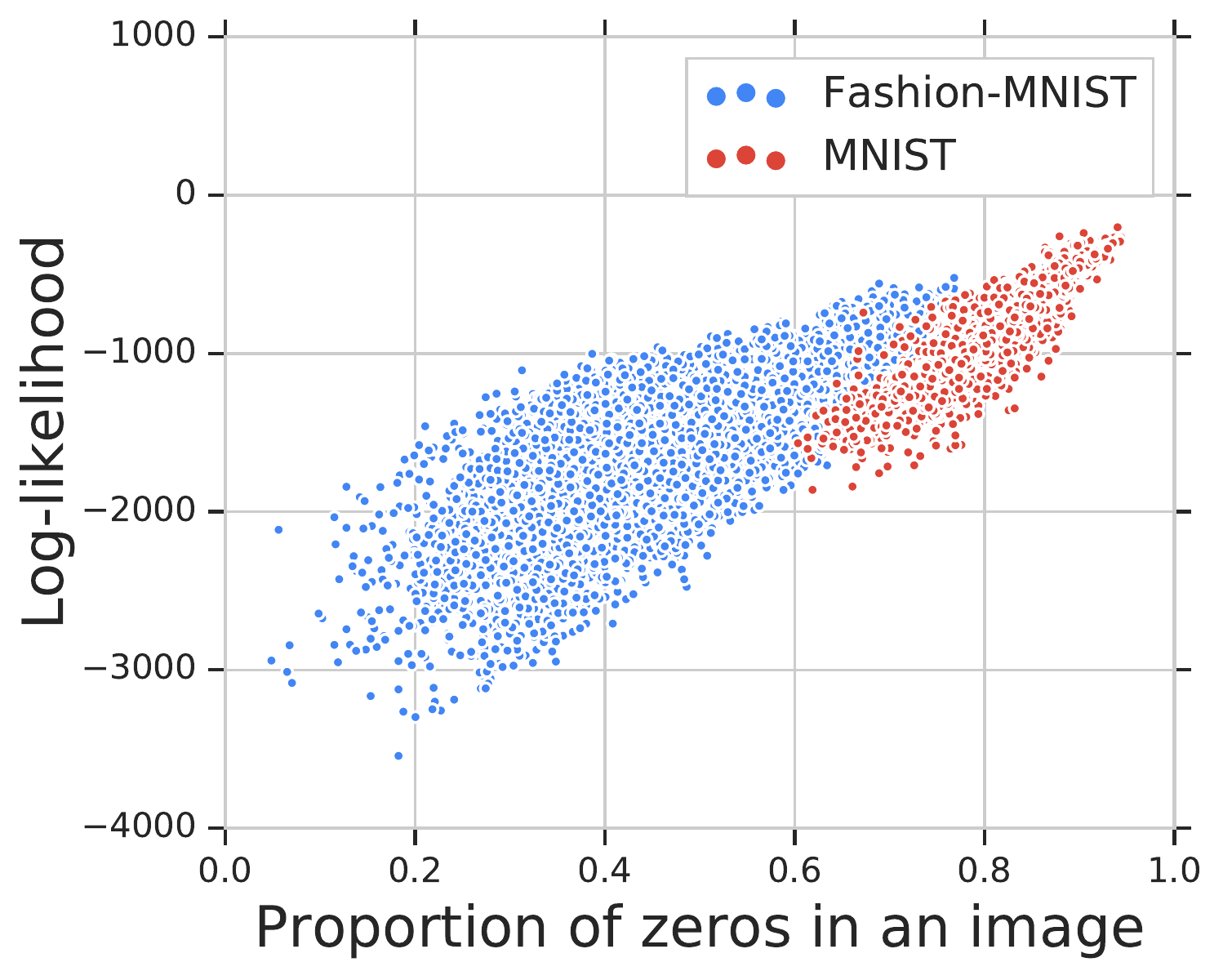} 
    }\end{subfigure}
    \begin{subfigure}[]{
    \includegraphics[width=0.325\textwidth]{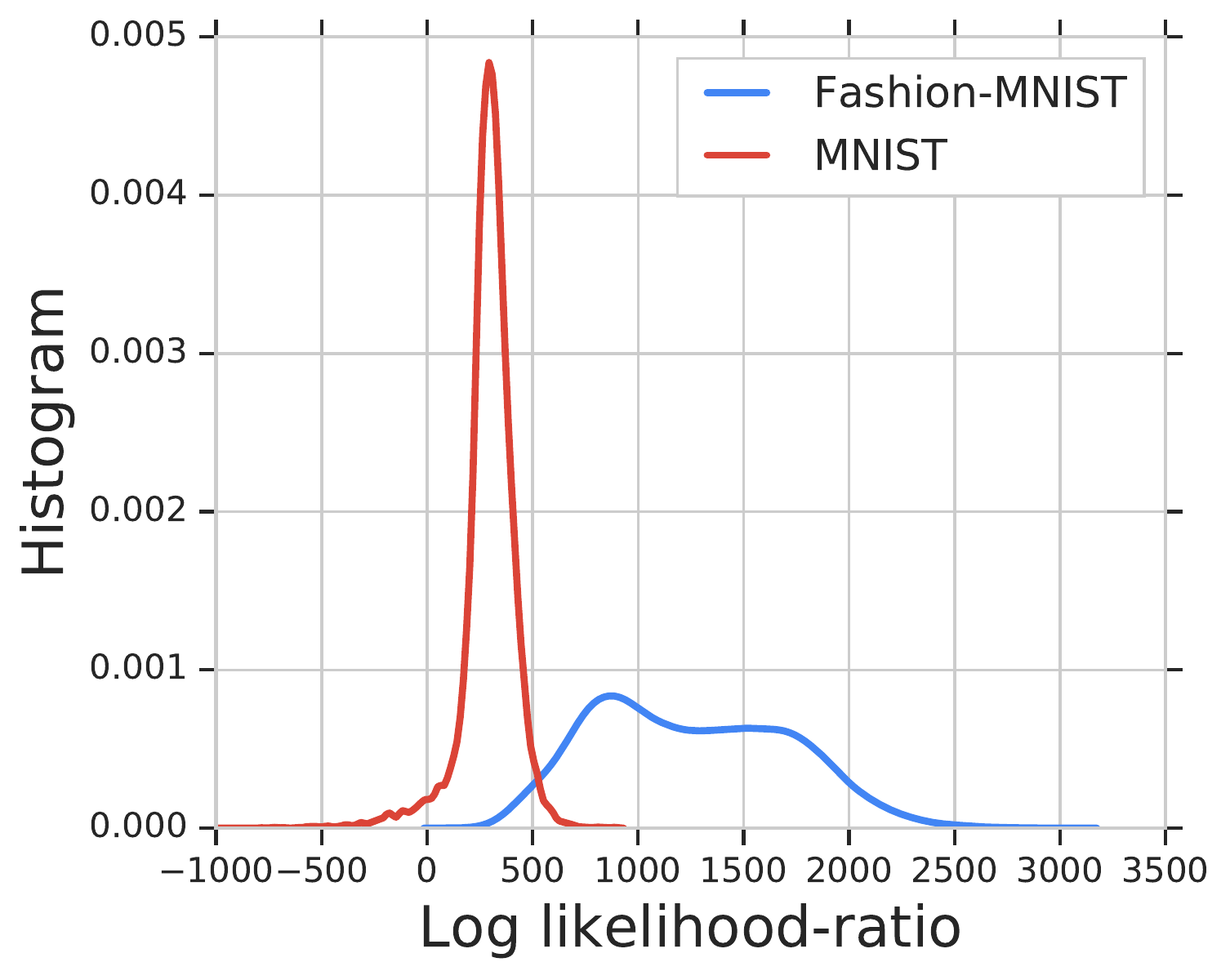}
    }\end{subfigure}
    \begin{subfigure}[]{
    \includegraphics[width=0.325\textwidth]{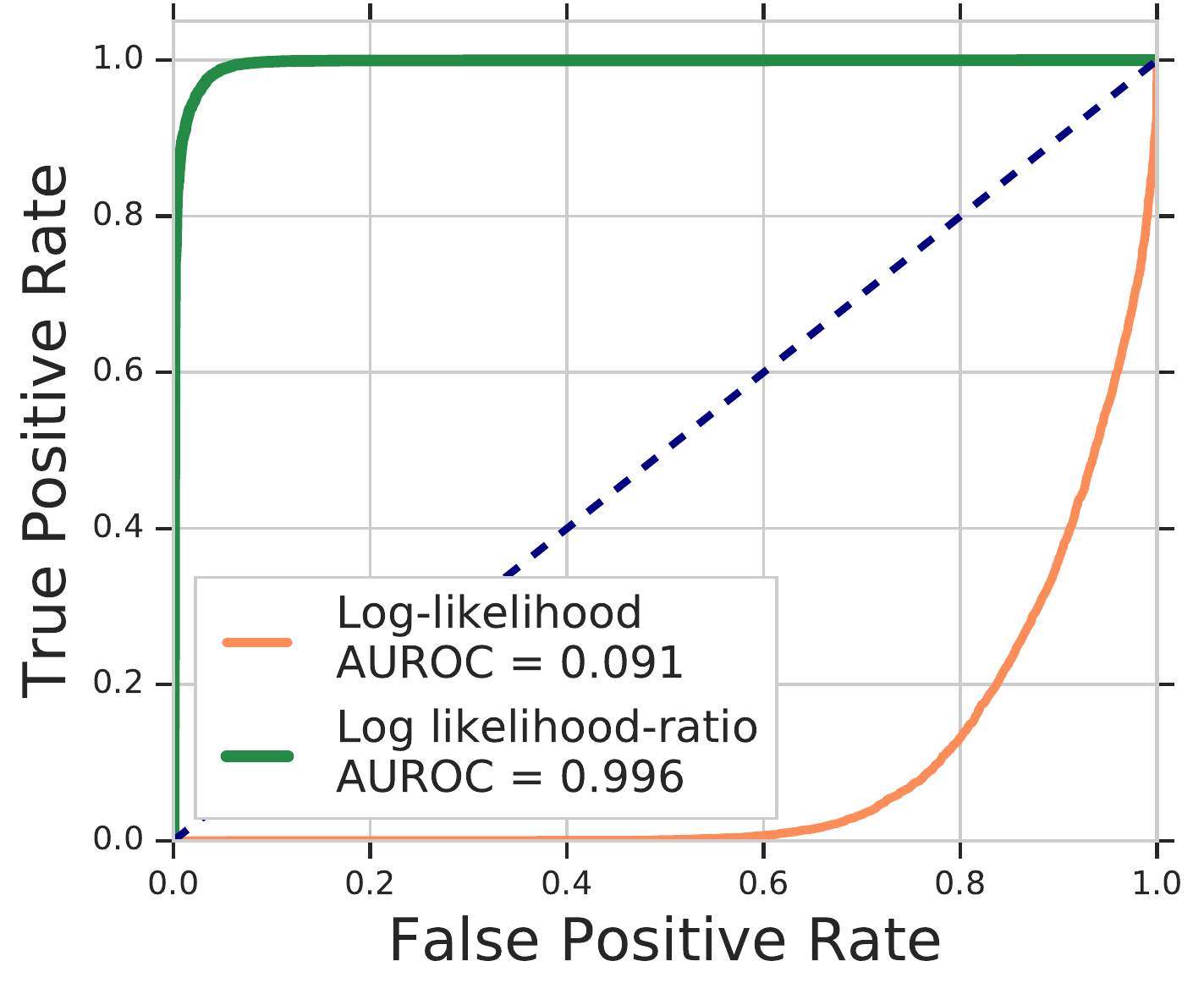}
    }\end{subfigure}
    \myvspace{-1em}
    \caption{(a) Log-likelihood of MNIST images (OOD) is higher than that of Fashion-MNIST images (in-distribution). (b) Log-likelihood is highly correlated with the background (proportion of zeros in an image). (c) Log-likelihood ratio is higher for Fashion-MNIST (in-dist) than MNIST (OOD). %
    (d) Likelihood ratio significantly improves the AUROC of OOD detection from 0.091 to 0.996. 
    }
    \label{fig:image_hist}
   \myvspace{-1em}
\end{figure}

Following existing literature \citep{nalisnick2018deep, hendrycks2018deep}, we evaluate our method using two experiments for detecting OOD images: (a) Fashion-MNIST as in-distribution and MNIST as OOD, (b) CIFAR-10 as in-distribution and SVHN as OOD.
For each experiment, we train a PixelCNN++ \citep{salimans2017pixelcnn++, van2016conditional} model using in-distribution data. %
We train a background model by adding perturbations to the training data. %
To compare with classifier-based baseline methods, we use CNN-based classifiers.
See Appendix \ref{sec:image_model} for model details. 
Based on the likelihood from the PixelCNN++ model, we confirm that the model assigns a higher likelihood to MNIST than Fashion-MNIST, as previously  reported by \citet{nalisnick2018deep}, and the AUROC for  OOD detection is only 0.091, even worse than  random (Figure~\ref{fig:image_hist}a). %
We discover that the proportion of zeros i.e.~\emph{number of pixels belonging to the background in an image is a confounding factor to the likelihood score} (Pearson Correlation Coefficient 0.85, see Figure~\ref{fig:image_hist}b, Figure~\ref{fig:showimage_likelihood}).
Taking the likelihood ratio between the original and the background models, we see that the AUROC improves significantly from 0.091 to 0.996 (Figure~\ref{fig:image_hist}d). 
The log likelihood-ratio for OOD images are highly concentrated around value 0, while that for in-distribution images are mostly positive (Figure~\ref{fig:image_hist}c). 

\begin{figure}[ht]%
    \centering
    \begin{subfigure}[Likelihood]{
        \includegraphics[width=0.22\textwidth]{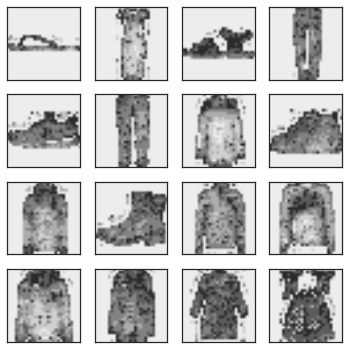}
    }\end{subfigure}
    \begin{subfigure}[Likelihood-Ratio]{
    \includegraphics[width=0.22\textwidth]{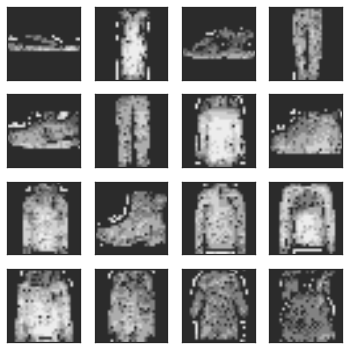}
    }\end{subfigure}
    \begin{subfigure}[Likelihood]{
    \includegraphics[width=0.22\textwidth]{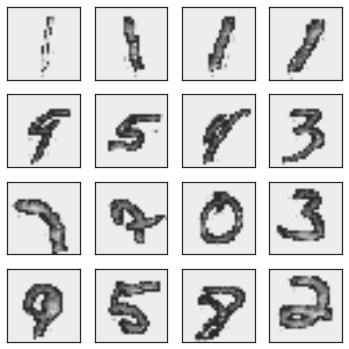}
    }\end{subfigure}
    \begin{subfigure}[Likelihood-Ratio]{
    \includegraphics[width=0.27\textwidth]{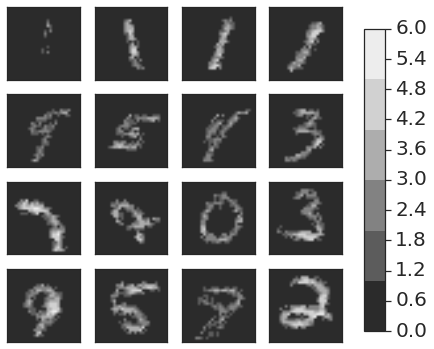} 
    }\end{subfigure}
    \vspace{-1em}
    \caption{The log-likelihood of each pixel in an image $\log p_{\vtheta}(x_d | \vx_{<d})$, and the log likelihood-ratio of each pixel $\log p_{\vtheta}(x_d | \vx_{<d}) - \log p_{\vtheta_0}(x_d | \vx_{<d}), d=1\dots, 784$., for 16 Fashion-MNIST images (a, b) and MNIST images (c, d).  Lighter gray color indicates larger value (see colorbar). Note that the range of log-likelihood (negative value) is different from that of log likelihood-ratio (mostly positive value). For the ease of visualization, we unify the colorbar by adding a constant to the log-likelihood score. The images are randomly sampled from the test dataset and sorted by their likelihood $p_{\theta}(\mathbf{x})$. Looking at which pixels contribute the most to each quantity, we observe that the likelihood value is dominated by the ``background'' pixels on both Fashion-MNIST and MNIST, whereas likelihood ratio focuses on the ``semantic'' pixels.}
    \label{fig:image_heat:mnist}
\end{figure}

\textbf{Which pixels contribute the most to the likelihood (ratio)?} To qualitatively evaluate the difference between the  likelihood and the likelihood ratio, we plot their values for each pixel for  Fashion-MNIST and MNIST images. This allows us to visualize which pixels contribute the most to the two terms respectively. 
Figure~\ref{fig:image_heat:mnist} shows a heatmap, with lighter (darker) gray colors indicating higher (lower) values.
Figures~\ref{fig:image_heat:mnist}(a,b) show that the likelihood value is dominated by the ``background'' pixels, whereas likelihood ratio focuses on the ``semantic'' pixels. Figures~\ref{fig:image_heat:mnist}(c,d) %
confirm that the background pixels cause MNIST images to be assigned high likelihood, whereas likelihood ratio focuses on the semantic pixels.   
We present additional qualitative results in Appendix~\ref{appendix:image}. For instance, Figure~\ref{fig:showimage_ratio} shows that images with the highest likelihood-ratios are those with prototypical Fashion-MNIST icons, e.g. ``shirts'' and ``bags'', highly contrastive with the background, while images with the lowest likelihood-ratios are those with rare patterns, e.g.~dress with stripes or sandals with high ropes.

We compare our method with other baselines. The classifier-based baseline methods are built using LeNet architecture. %
Table \ref{tab:both}a
shows that our method achieves the highest AUROC$\uparrow$, AUPRC$\uparrow$, and the lowest FPR80$\downarrow$.  
The method using Mahalanobis distance performs better than other baselines.
Note that the binary classifier between in-distribution and perturbed in-distribution does not perform as well as our method, possibly due to the fact that while the features learned by the discriminator can be good for detecting perturbed inputs, they  
may not generalize well for OOD detection. 
The generative model approach based on $p(\vx)$ captures more fundamental features of the data generation process than the discriminative approach. 

\begin{table}
\caption{AUROC$\uparrow$, AUPRC$\uparrow$, and FPR80$\downarrow$ for detecting OOD inputs using likelihood and likelihood-ratio method and other baselines on (a) Fashion-MNIST vs. MNIST datasets and (b) genomic dataset. The up and down arrows on the metric names indicate whether greater or smaller is better. 
$\mu$ in the parentheses indicates the background model is tuned only using noise perturbed input, and ($\mu$ and $\lambda$) indicates the background model is tuned by both perturbation and $L_2$ regularization. Numbers in front and inside of the brackets are mean and standard error respectively based on 10 independent runs with random initialization of network parameters and random shuffling of training inputs. For ensemble models, the mean and standard error are estimated based on 10 bootstrap samples from 30 independent runs, which can be underestimations of the true standard errors.}
\vspace{-0.5em}
\begin{minipage}[b]{0.50\linewidth}
\myvspace{-2em}
\begin{adjustbox}{max width=\textwidth}
\centering
\begin{tabular}{lccc}%
 &AUROC$\uparrow$&AUPRC$\uparrow$&FPR80$\downarrow$  \\ \hline
Likelihood                       & 0.089 (0.002) & 0.320 (0.000) & 1.000 (0.001)  \\
Likelihood Ratio (ours, $\mu$) &  0.973 (0.031) & 0.951 (0.063) & 0.005 (0.008) \\
Likelihood Ratio (ours, $\mu, \lambda$) & \textbf{0.994 (0.001)} & \textbf{0.993 (0.002)} & \textbf{0.001 (0.000)} \\
$p(\hat{y}|\vx)$               & 0.734 (0.028) & 0.702 (0.026) & 0.506 (0.046)    \\
Entropy of $p(y|\vx)$          & 0.746 (0.027) & 0.726 (0.026) & 0.448 (0.049)    \\
ODIN                                  & 0.752 (0.069) & 0.763 (0.062) & 0.432 (0.116) \\
Mahalanobis distance                  & 0.942 (0.017) & 0.928 (0.021) & 0.088 (0.028)  \\
Ensemble, 5 classifiers & 0.839 (0.010) & 0.833 (0.009) & 0.275 (0.019) \\
Ensemble, 10 classifiers & 0.851 (0.007) & 0.844 (0.006) & 0.241 (0.014) \\
Ensemble, 20 classifiers & 0.857 (0.005) & 0.849 (0.004) & 0.240 (0.011)  \\ 
Binary classifier                     & 0.455 (0.105) & 0.505 (0.064) & 0.886 (0.126) \\
$p(\hat{y}|\vx)$ with noise class & 0.877 (0.050) & 0.871 (0.054) & 0.195 (0.101) \\
$p(\hat{y}|\vx)$ with calibrations & 0.904 (0.023) & 0.895 (0.023) & 0.139 (0.044) \\  
WAIC, 5 models                  & 0.221 (0.013) & 0.401 (0.008) & 0.911 (0.008) \\ \hline \\
\end{tabular}
\end{adjustbox}
\centerline{(a)}
\end{minipage}
\hfill
\begin{minipage}[b]{0.50\linewidth}
\centering
\begin{adjustbox}{max width=\textwidth}
\begin{tabular}{lccc}%
 & AUROC$\uparrow$ & AUPRC$\uparrow$ & FPR80$\downarrow$ \\ \hline
Likelihood                & 0.626 (0.001) & 0.613 (0.001) & 0.661 (0.002) \\
Likelihood Ratio (ours, $\mu$)  & 0.732 (0.015) & 0.685 (0.017) & 0.534 (0.031) \\
Likelihood Ratio (ours, $\mu, \lambda$)  & \textbf{0.755 (0.005)} & \textbf{0.719 (0.006)} & \textbf{0.474 (0.011)} \\
 $p(\hat{y}|\vx)$                 & 0.634 (0.003) & 0.599 (0.003) & 0.669 (0.007) \\
Entropy of $p(y|\vx)$             & 0.634 (0.003) & 0.599 (0.003) & 0.617 (0.007) \\
Adjusted ODIN                     & 0.697 (0.010) & 0.671 (0.012) & 0.550 (0.021) \\
Mahalanobis distance              & 0.525 (0.010) & 0.503 (0.007) & 0.747 (0.014) \\
Ensemble, 5 classifiers           & 0.682 (0.002) & 0.647 (0.002) & 0.589 (0.004) \\
Ensemble, 10 classifiers          & 0.690 (0.001) & 0.655 (0.002) & 0.574 (0.004) \\
Ensemble, 20 classifiers          & 0.695 (0.001) & 0.659 (0.001) & 0.570 (0.004) \\
Binary classifier                 & 0.635 (0.016) & 0.634 (0.015) & 0.619 (0.025) \\
$p(\hat{y}|\vx)$ with noise class & 0.652 (0.004) & 0.627 (0.005) & 0.643 (0.008) \\
$p(\hat{y}|\vx)$ with calibration & 0.669 (0.005) & 0.635 (0.004) & 0.627 (0.006) \\
WAIC, 5 models                    & 0.628 (0.001) & 0.616 (0.001) & 0.657 (0.002) \\ \hline \\
\end{tabular}
\end{adjustbox}
\centerline{(b)}
\end{minipage}
\myvspace{-1em}
\label{tab:both}
\end{table}

For the experiment using CIFAR-10 as in-distribution and SVHN as OOD, we apply the same training procedure using the PixelCNN++ model and choose hyperparameters using grayscaled CIFAR-10 which was shown to be OOD by \citet{nalisnick2018deep}. See Appendix~\ref{sec:image_model} for model details.  
Looking at the results in Table~\ref{tab:cifar-llr}, 
we observe that the OOD images from  SVHN  have  higher likelihood than the in-distribution images from CIFAR-10, confirming the observations of \citet{nalisnick2018deep}, with AUROC of 0.095. 
Our likelihood-ratio method  significantly improves the  AUROC to 0.931. Figure~\ref{fig:image_heat:cifar} in Appendix~\ref{appendix:image} shows additonal qualitative results. 
For detailed results including other baseline methods, see Table \ref{tab:cifar} in Appendix~\ref{sec:image_table}. 
\begin{table}
\centering
\caption{
CIFAR-10 vs SVHN results:  
AUROC$\uparrow$, AUPRC$\uparrow$, FPR80$\downarrow$ for detecting OOD inputs using likelihood and our likelihood-ratio method. %
}
\vspace{-0.5em}
\begin{adjustbox}{max width=0.6\textwidth}
\begin{tabular}{llll}
& AUROC$\uparrow$ & AUPRC$\uparrow$ & FPR80$\downarrow$       \\ \hline
Likelihood & 0.095 (0.003) &	0.320 (0.001) & 1.000 (0.000) \\
Likelihood Ratio (ours, $\mu$) & 0.931 (0.032)	& 0.888 (0.049) & 0.062 (0.073) \\
Likelihood Ratio (ours, $\mu$, $\lambda$) & 0.930	(0.042)	& 0.881	(0.064)	& 0.066	(0.123) \\ \hline
\end{tabular}
\end{adjustbox}
\myvspace{-2em} 
\label{tab:cifar-llr}
\end{table}

\subsection{OOD detection for genomic sequences}\label{sec:genomics}

\textbf{Dataset for detecting OOD genomic sequences}
\begin{figure}[ht]
\begin{center}
    \centering
    \includegraphics[width=0.48\textwidth]{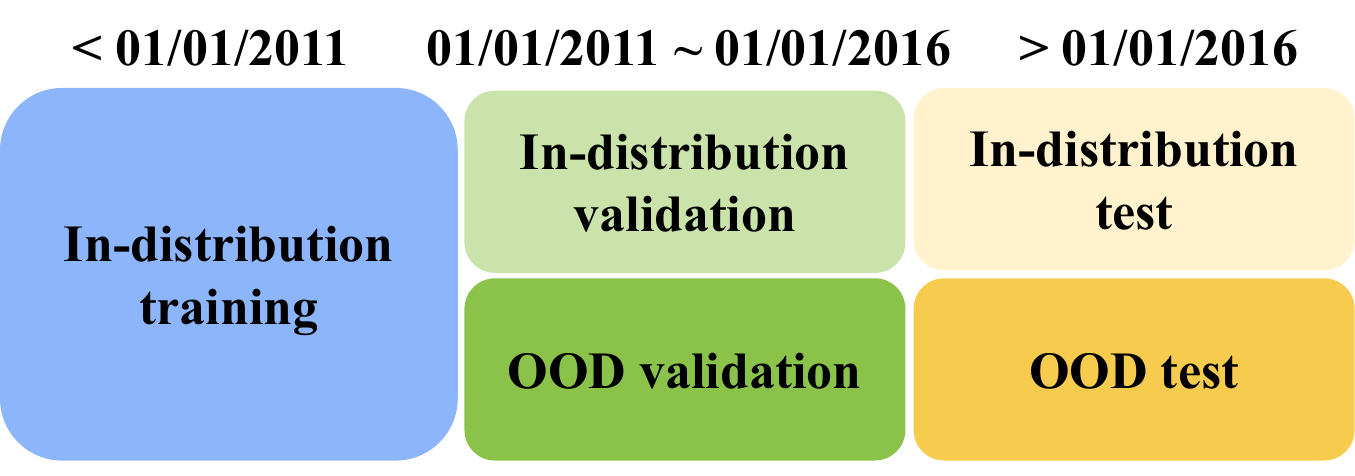}
    \caption{The design of the training, validation, and test datasets for genomic sequence classification including in and OOD data. 
    }
    \label{fig:dataset}
\end{center}
\end{figure}
We design a new dataset for evaluating OOD methods. 
As bacterial classes are discovered gradually over time, in- and out-of-distribution data can be naturally separated by the time of discovery.  
Classes discovered before a cutoff time can be regarded as in-distribution classes, and those discovered afterward, which were unidentified at the cutoff time, can be regarded as OOD. 
We choose two cutoff years, 2011 and 2016, to define the training, validation, and test splits (Figure~\ref{fig:dataset}).
Our dataset contains of 10 in-distribution classes, 60 OOD classes for validation, and 60 OOD classes for testing.
Note that the validation OOD dataset is only used for hyperparameter tuning, {and the validation OOD classes are disjoint from the test OOD classes}.
To mimic sequencing data, we fragmented genomes in each class into short sequences of 250 base pairs, which is a common length that current sequencing technology generates. 
Among all the short sequences, we randomly choose 100,000 sequences for each class for training, validation, and test.
Additional details about the dataset, including pre-processing and the information for the in- and out-of-distribution classes, can be found in Appendix \ref{dataset}.

\textbf{Likelihood ratio method for detecting OOD sequences}
We build an LSTM model for estimating the likelihood $p(\vx)$ based on the transition probabilities $p(x_d|\vx_{<d})$, $d=1,\dots, D$. 
In particular, we feed the one-hot encoded DNA sequences into an LSTM layer, followed by a dense layer and a softmax function to predict the probability distribution over the 4 letters of $\{A, C, G, T\}$, and train the model using only the in-distribution training data.
We evaluate the likelihood for sequences in the OOD test dataset under the trained model, and compare those with the likelihood for sequences in the in-distribution test dataset. 
The AUROC$\uparrow$, AUPRC$\uparrow$, and FPR80$\downarrow$ scores are 0.626, 0.613, and 0.661 respectively 
(Table 
\ref{tab:both}b).

\begin{figure}[ht]
     \centering
    \begin{subfigure}[]{
        \includegraphics[width=0.31\textwidth]{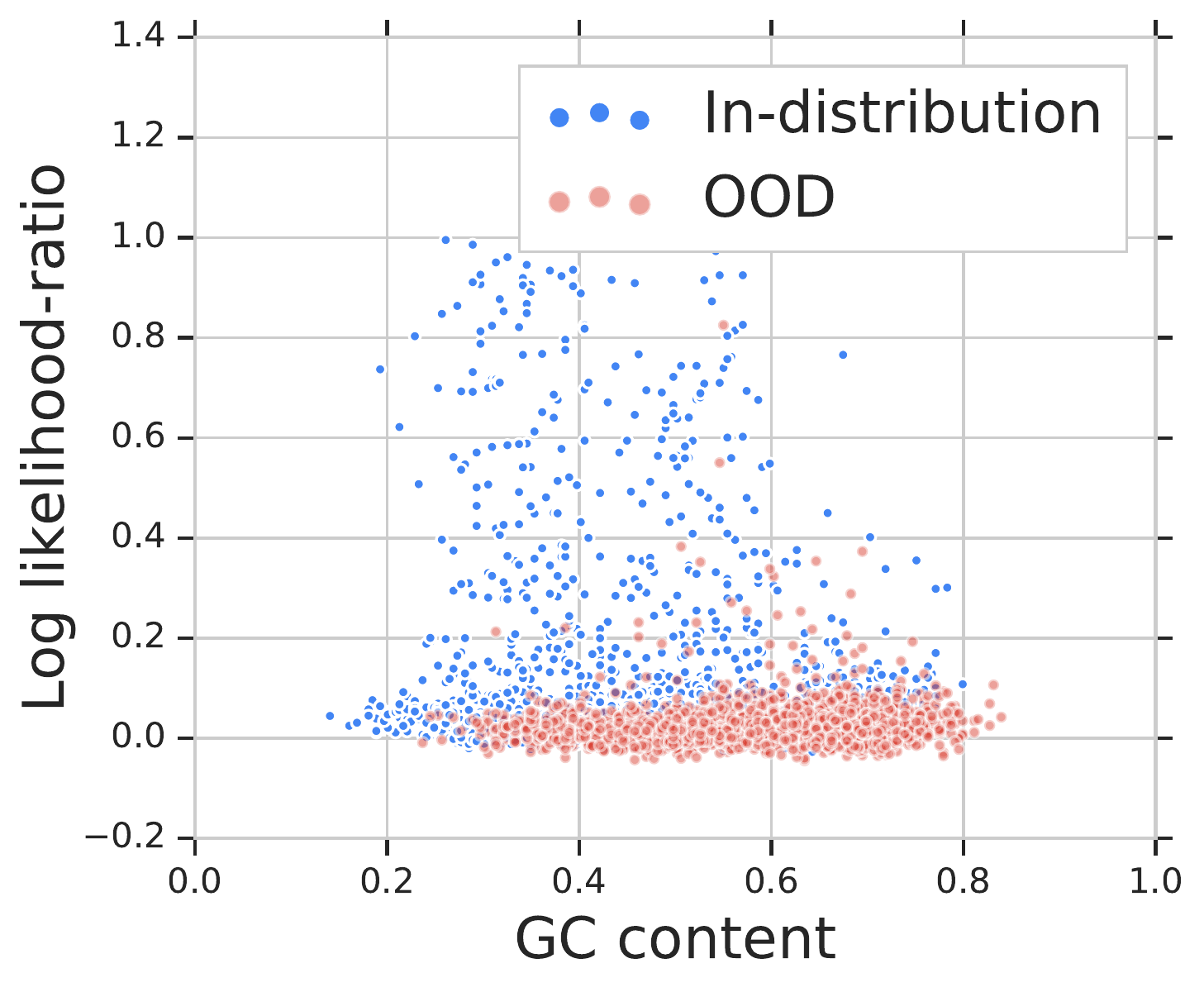} %
    }\end{subfigure}
    \begin{subfigure}[]{
        \includegraphics[width=0.31\textwidth]{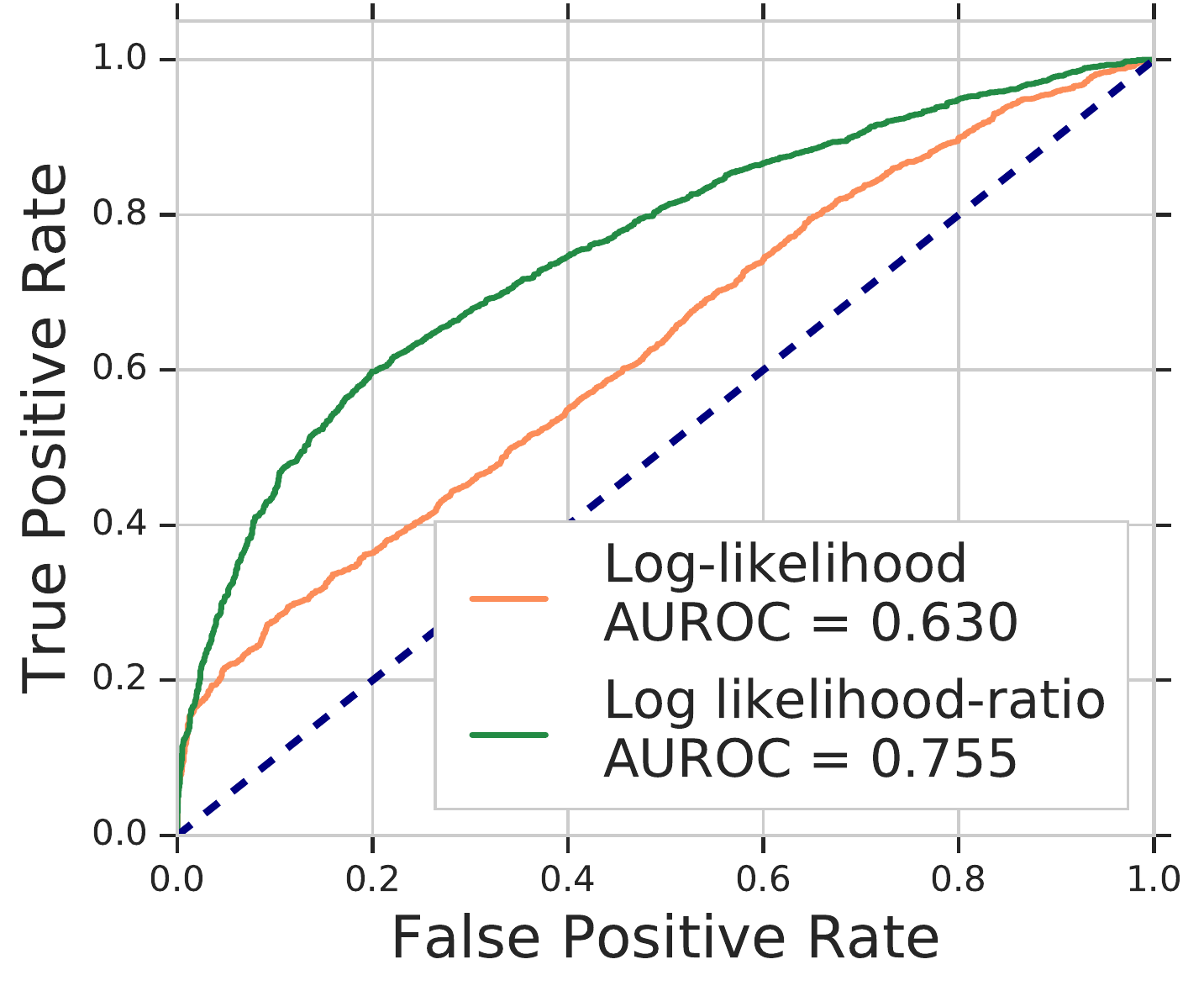} %
    }\end{subfigure}
    \begin{subfigure}[]{
        \includegraphics[width=0.31\textwidth]{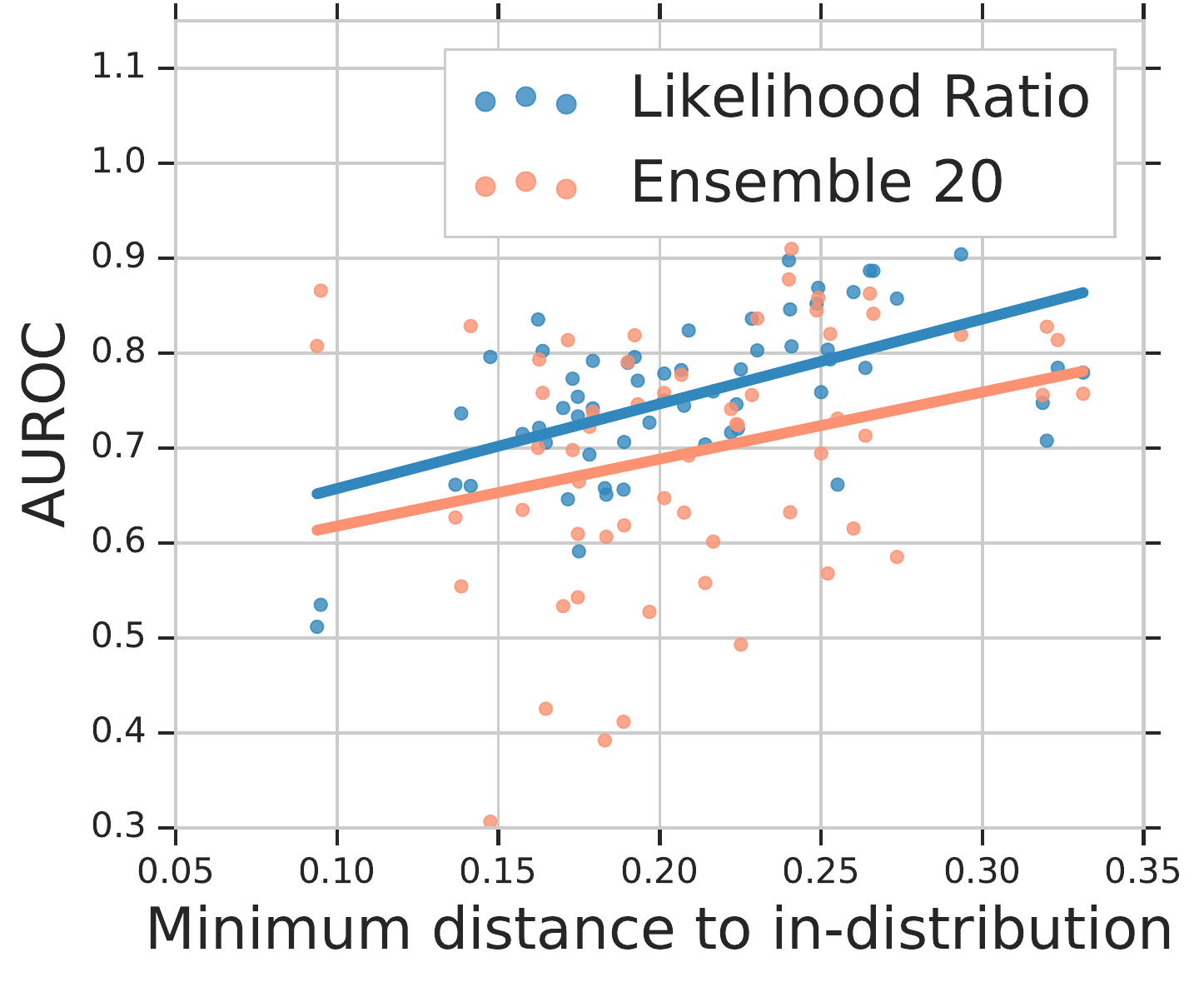} %
    }\end{subfigure}
    \myvspace{-1em}
    \caption{
    (a) The  likelihood-ratio score is roughly independent of the  GC-content  which makes it less susceptible to background statistics and better suited for OOD detection. %
    (b) ROCs and AUROCs for OOD detection using likelihood and likelihood-ratio. (c) Correlation between the AUROC of OOD detection and distance to in-distribution classes using Likelihood Ratio and the Ensemble method. %
    }
        \myvspace{-0.5em}
    \label{fig:genomics}
\end{figure}

We  train a background model by using the perturbed in-distribution data and optionally adding $L_2$ regularization to the model weights. Hyperparameters are tuned using validation dataset which contains in-distribution and validation OOD classes, and the validation OOD classes are disjoint from test OOD classes.   
Contrasting against the background model, the AUROC$\uparrow$, AUPRC$\uparrow$, and FPR80$\downarrow$ for the likelihood-ratio significantly improve to 0.755, 0.719, and 0.474, respectively 
(Table \ref{tab:both}b,
Figure~\ref{fig:genomics}b). 
Compared with the likelihood, the AUROC and AUPRC for likelihood-ratio increased 20\% and 17\% respectively, and the FPR80 decreased 28\%. 
Furthermore, Figure~\ref{fig:genomics}a shows that the likelihood ratio is less sensitive to GC-content, and the separation between in-distribution and OOD distribution becomes clearer. %
We evaluate other baseline methods on the test dataset as well. 
For classifier-based baselines, we construct CNNs with one convolutional layer, one max-pooling layer, one dense layer, and a final dense layer with softmax activation for predicting class probabilities, as in \citet{alipanahi2015predicting, busia2018deep, ren2018identifying}. 
Comparing our method to the baselines in 
Table~\ref{tab:both}b,
our method achieves the highest AUROC, AUPRC, and the lowest FPR80 scores on the test dataset. 
Ensemble method and ODIN perform better than other baseline methods. 
Comparing with the Fashion-MNIST and MNIST experiment, the Mahalanobis distance performs worse for detecting genomic OOD possibly due to the fact that Fashion-MNIST and MNIST images are quite distinct while in-distribution and OOD bacteria classes are interlaced under the same taxonomy (See Figure \ref{fig:phylo} for the phylogenetic tree of the in-distribution and OOD classes). 

\textbf{OOD detection correlates with its distance to in-distribution} %
We investigate the effect of the distance between the OOD class to the in-distribution classes, %
on the performance of OOD detection.  
To measure the distance between the OOD class to the in-distribution, we randomly select  representative genome from each of the in-distribution classes and OOD classes. 
We use the state-of-the-art alignment-free method for genome comparison, $d_2^S$ \citep{ren2018alignment, reinert2009alignment}, to compute the genetic distance between each pair of the genomes in the set.  
This genetic distance is calculated based on the similarity between the normalized nucleotide word frequencies ($k$-tuples) of the two genomes, and studies have shown that this genetic distance reflects true evolutionary distances between genomes \citep{chan2014inferring, bernard2016alignment, lu2017cafe}.
For each of the OOD classes, we use the minimum distance between the genome in that class to all genomes in the in-distribution classes as the measure of the genetic distance between this OOD class and the in-distribution. 
Not surprisingly, the the AUROC for OOD detection is positively  correlated with the genetic distance (Figure~\ref{fig:genomics}c), and an OOD class far away from in-distribution is easier to be detected. 
Comparing our likelihood ratio method and one of the best classifier-based methods, ensemble method, we observe that our likelihood ratio method has higher AUROC for different OOD classes than ensemble method in general. 
Furthermore, our method has a higher Pearson correlation coefficient (PCC) of 0.570 between the minimum distance and AUROC for Likelihood Ratio method, than the classifier-based ensemble method with 20 models which has PCC of 0.277. 
The dataset and code for the genomics study is available at \url{https://github.com/google-research/google-research/tree/master/genomics_ood}.

  \myvspace{-0.5em}
\section{Discussion and Conclusion}
  \myvspace{-0.5em}

We investigate deep generative model-based methods for OOD detection and show that the likelihood of auto-regressive models can be confounded by %
background statistics, providing an explanation to the failure of PixelCNN for OOD detection observed by recent work  \citep{nalisnick2018deep,hendrycks2018deep,shafaei2018does}. %
We propose a likelihood ratio method that alleviates this issue by contrasting the likelihood against a background model.  
We show that our method effectively corrects %
for the background components, and significantly improves the accuracy of OOD detection on both image datasets and genomic datasets. %
Finally, we create and 
release a realistic genomic sequence dataset for OOD detection which highlights an important real-world problem, and 
hope that this serves as a valuable OOD detection benchmark for the research community.  

\subsubsection*{Acknowledgments}
We thank 
Alexander A. Alemi, 
Andreea Gane,
Brian Lee,
D. Sculley,
Eric Jang, 
Jacob Burnim,
Katherine Lee,
Matthew D. Hoffman, 
Noah Fiedel, 
Rif A. Saurous,
Suman Ravuri,
Thomas Colthurst,
Yaniv Ovadia,
the Google Brain Genomics team, and Google TensorFlow Probability team 
for helpful feedback and discussions.

\bibliography{main}
\bibliographystyle{icml2019}
\clearpage
\newpage
\appendix

\setcounter{figure}{0}
\setcounter{table}{0}
\makeatletter 
\renewcommand{\thefigure}{S\@arabic\c@figure}
\renewcommand{\thetable}{S\@arabic\c@table}
\makeatother

{\Large{\textbf{
\begin{center}
Appendix  
\end{center}
}}}
\section{Additional details about our proposed likelihood ratio  method}\label{sec:pseudocode}

The pseudocode for generating input perturbations for training the background model is described in Algorithm \ref{alg:noise}.
\begin{algorithm}[H]%
   \caption{%
   Input perturbation for background model training}
   \label{alg:noise}
\begin{algorithmic}[1]
  \STATE \textbf{Inputs}: 
 $D$-dimensional input $\vx = x_1\dots x_D$, Mutation rate $\mu$ and vocabulary $\mathcal{A}$. Note that we assume inputs to be discrete, i.e.~$x_d\in \mathcal{A}$, where $\mathcal{A}=\{A,C,G,T\}$ for genomic sequences and $\mathcal{A}=\{0,\dots,255\}$ for images.
 \STATE \textbf{Output}: perturbed input $\tilde{\vx}$
 \STATE Generate a $D$-dimensional vector $\mathbf{v}=v_1\dots,v_D$, where $v_d \in \{0, 1\}$ are independent and identically distributed according to a  Bernoulli distribution with rate $\mu$.
  \FOR{index $d \in \{1, \ldots, D\}$} 
 \IF{$v_d=1$}
 \STATE Sample $\tilde{x}_d$ from the set $\mathcal{A}$ with equal probability. 
 \ELSE \STATE Set $\tilde{x}_d = x_d$. %
 \ENDIF
  \ENDFOR
\end{algorithmic}
\end{algorithm}

The complete pseudocode for our method is described in Algorithm~\ref{alg:llr}. 
The runtime of our method is two times of the standard generative model runtime.
\begin{algorithm}[H]%
   \caption{%
   OOD detection using Likelihood Ratio}
   \label{alg:llr}
\begin{algorithmic}[1]
  \STATE Fit a %
  model $p_{\vtheta}(\vx)$ using %
  in-distribution dataset $\mathcal{D}_{\text{in}}$.
   \STATE Fit a background model $p_{\vtheta_0}(\vx)$ using perturbed input data $\widetilde{\mathcal{D}}_{\text{in}}$  (generated using Algorithm~\ref{alg:noise}) and (optionally)  model regularization techniques.  
   \STATE Compute the likelihood ratio statistic $\llr(\vx)$ %
   \STATE Predict OOD if $\llr(\vx)$ is small.
\end{algorithmic}
\end{algorithm}
\section{Supplementary materials for the experiments on images}
\label{appendix:image}
\subsection{Model details} \label{sec:image_model}
Following existing literature \citep{nalisnick2018deep, choi2018waic}, we evaluate our method using two experiments for detecting OOD images: (a) Fashion-MNIST as in-distribution and MNIST as OOD, (b) CIFAR-10 as in-distribution and SVHN as OOD.
For the experiment of Fashion-MNIST vs. MNIST, we train a generative model using the training set of Fashion-MNIST, and use the test set of Fashion-MNIST and the test set of MNIST as OOD as the final test dataset. 
The same rule is applied for CIFAR-10 vs. SVHN experiment. Since SVHN test set has more inputs than CIFAR-10 test set, we randomly select the same number of inputs for evalaution.

For each experiment, we train a PixelCNN++ \citep{salimans2017pixelcnn++, van2016conditional} model on the  in-distribution data using maximum likelihood. %
For Fashion-MNIST dataset, the model uses 2 blocks of 5 gated ResNet layers with 32 convolutional 2D filters (concatenate ELU activation function, and without weight normalization and dropout), and 1 component in the logitstic mixture, and is trained for 50,000 steps with initial learning rate of 0.0001 with exponential decay rate of 0.999995 per step, batch size of 32, and Adam optimizer with momentum parameter $\beta_1=0.95$ and $\beta_2=0.9995$. 
For CIFAR-10 dataset, the model uses 2 blocks of 5 gated ResNet layers with 160 convolutional 2D filters, and 10 components in the logitstic mixture, and is trained for 600,000 steps with the same learning rate, batch size, and optimizer as the above. 
The bits per dimension of the PixelCNN++ models are 2.92 and 1.64 for in-distribution images Fashion-MNIST and OOD images MNIST respectively, and
3.20 and 2.15 for in-distribution images CIFAR-10 and OOD images SVHN, respectively.

For the background model, we train a PixelCNN++ model with the same architecture on perturbed inputs obtained by randomly flipping input pixel values to one of the 256 possible values following an independent and identical Bernoulli distribution with rate $\mu$ (see Algorithm \ref{alg:noise}). 
The mutation rate $\mu$ for adding perturbations to input in the background model training is a hyperparameter to our method.
For tuning the hyperparameters, we use independent datasets, NotMNIST \citep{notmnist} for Fashion-MNIST vs. MNIST experiment, and gray-scaled CIFAR-10 for CIFAR-10 vs. SVHN experiment, as the validation OOD dataset. 
We choose the optimal mutation rate $\mu$ based on the AUROC for OOD detection using the validation dataset.
We test the mutation rate $\mu$ from the range of $[0.1, 0.2, 0.3]$. 
For Fashion-MNIST vs. MNIST, the optimal mutation rate tuned using NotMNIST as OOD dataset is $\mu=0.3$ for most of the 10 independent runs.
We also add $L_2$ regularization to model weights. We let the $L_2$ coefficient $\lambda$ ranges from $\lambda=[0, 10, 100]$ and test different combinations of $\mu$ and $\lambda$ on the background model training. 
We find that adding $L_2$ regularization helps to improve the final test AUROC only slightly from 0.973 to 0.996.
The optimal combination is $\mu=0.3$ and $\lambda=100$ for most of the 10 independent runs of random initialization of network parameters and random shuffling of training inputs. 
Table \ref{tab:hyper_image}a shows one of those, which results in AUROC of 0.996 in the final test dataset. 
For CIFAR-10 vs. SVHN experiment, we observe that tuning for both mutation rate and  $L_2$ regularization achieves similar performance with tuning for mutation rate only. 
The optimal mutation rate is $\mu=0.1$ for most of the 10 independent runs (data not shown).

In the case where no independent OOD data (such as NotMNIST) is available for hyperparameter tuning, we can use randomly mutated in-distribution input at mutation rate 2\%, to mimic the OOD input. 
The mutation is added using the same procedure as that for training the background model.
The optimal hyperparameter setting is $\mu=0.1$ and $\lambda=100$, which achieves AUROC of 0.958 in the final test dataset.
The results show that the hyperparameters for the background model are easy to tune.
Under the situation where indepedent OOD data are not available, using only simulated OOD data we are able to achieve reasonable performance. 

We found that the performance of our method LLR can be slightly affected by different PixelCNN++  network setups. 
We tested both versions of PixelCNN++ where input images ranging from 0 to 255 are (a) directly fed into the networks (b) re-scaled to the range of -1 to 1 and fed into the networks. 
The two setups give different initializatiobs of the network.
We found that the version without rescaling achieves slightly better performance in terms of AUROC possibly because it learns a better background model based on its initialization. 
The AUROC$\uparrow$, AUPRC$\uparrow$, and FPR80$\downarrow$ for Fashion-MNIST vs. MNIST in Table \ref{tab:both}a 
are based on the version without rescaling. 
Using the version with rescaling, the numbers for the three evaluation metrics are  0.936 (0.003), 0.891 (0.004), 0.025 (0.003), without changing model parameters (the same 5 gated ResNet with 32 convolutional 2D filters, batch size, learning rate, optimizer as before), and training for 600,000 steps.
For CIFAR-10 vs. SVHN experiment, we found rescaling helps to produce more stable results. So we report AUROC$\uparrow$, AUPRC$\uparrow$, and FPR80$\downarrow$ based on the version with rescaling in Table \ref{tab:cifar}.

\begin{table}[ht]
\caption{Hyperparameter tuning of mutation rate $\mu$ and $L_2$ coefficient $\lambda$
of the background model of our likelihood ratio method for Fashion-MNIST vs. MNIST experiment.  %
(a) 
AUROC is evaluated based on in-distribution Fashion-MNIST validation dataset and NotMNIST dataset. Note that MNIST is not used for hyperparameter turning. 
(b) The same as (a) but tuning using simulated OOD inputs without exposing to any NotMNIST or MNIST images. The simulated OOD inputs are generated by permuting the in-distribution inputs at the mutation rate 2\%. 
}
\hfill
\begin{minipage}[b]{0.48\linewidth}
\centering
\begin{tabular}{clllll}
& $\mu=0.1$ & 0.2   & 0.3   &       \\ \hline
$\lambda=0$   & 0.358 & 0.426 & \textbf{0.795} \\
10     & 0.433 & 0.432 & 0.489 \\
100    & 0.414 & 0.777 & \textbf{0.798} \\ \hline
\end{tabular}
\end{minipage}
\hfill
\begin{minipage}[b]{0.48\linewidth}
\centering
\begin{tabular}{clllll}
& $\mu=0.1$ & 0.2   & 0.3   &       \\ \hline
$\lambda=0$   & \textbf{0.984} & 0.971 & 0.856 \\
10     & \textbf{0.989} & 0.977 & 0.971 \\
100    & \textbf{0.989} & 0.851 & 0.857 \\ \hline
\end{tabular}
\end{minipage}
\myvspace{1em}
\label{tab:hyper_image}
\end{table}

To compare with classifier-based baseline methods, we build convolutional neural networks (CNNs). We used LeNet \citep{lecun1998gradient} architecture for Fashion-MNIST images. The model composes two convolutional layers with 32 and 64 2D filters respectively of size 3 by 3 with ReLU activation function, a max pooling layers of size 2 by 2, a dropout layer with the rate of 0.25, a dense layer of 128 units with ReLU activation function, another dropout layer with the rate of 0.25 based on the flattened output from the previous layer, and a final dense layer with the softmax activation function used for generating the class probabilities. 
Model parameters are trained for 12 epochs with batch size of 128, learning rate of 0.002, and Adam optimizer. 
The prediction accuracy on test data is 0.923 for in-distribution Fashion-MNIST images.
For CIFAR-10 images, we build ResNet-20 V1 architecture with ReLU activations \citep{he2016deep}.
Model parameters are trained for 120 epochs with batch size of 7, Adam optimizer, and a learning rate schedule that multiplied an initial learning rate of 0.0007 by 0.1, 0.01, 0.001, and 0.0005 at steps 80, 120, 160, and 180 respectively. Training inputs are randomly distorted using horizontal flips and random crops preceded by 4-pixel padding as described in \cite{he2016deep}. The prediction accuracy on test data is 0.911 for in-distribution CIFAR-10 images. 
Both LeNet and ResNet-20 V1 model architectures are consistent with those in \cite{ovadia2019can}. 

For the baseline methods 6-8 that are based on perturbed inputs, the perturbation rate is tuned from the range of $[10^{-5}, 10^{-4}, \dots, 10^{-1}]$ based on validation in-distribution dataset and an independent dataset that is different from the final OOD test dataset, e.g. NotMNIST for Fashion-MNIST vs. MNIST experiment, and grayscaled CIFAR-10 for CIFAR-10 vs. SVHN experiment.
Similarly the hyperparameters in ODIN method, the temperature scaling to logits and perturbations to inputs, are tuned based on the same validation dataset.
The temperature is tuned in the range of $[1, 5, 10, 100, 1000]$, and the perturbation is tuned in the range of $[0, 10^{-8}, 10^{-7}, \dots, 10^{-1}]$.

\subsection{Supplementary figures}

\begin{figure}[ht]%
    \centering
     \begin{subfigure}[FashionMNIST: highest log-likelihood.]{
     \includegraphics[width=0.22\textwidth]{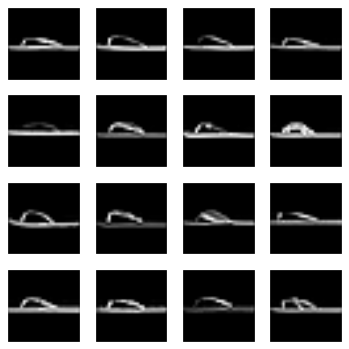}
     }\end{subfigure}
     \hfill
      \begin{subfigure}[FashionMNIST: lowest log-likelihood.]{
    \includegraphics[width=0.22\textwidth]{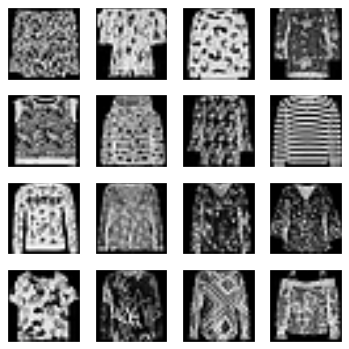} 
      }\end{subfigure}
      \hfill 
      \begin{subfigure}[MNIST: highest log-likelihood.]{
    \includegraphics[width=0.22\textwidth]{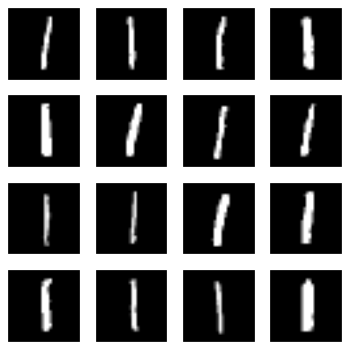}
      }\end{subfigure}
       \hfill
      \begin{subfigure}[MNIST: lowest log-likelihood.]{
    \includegraphics[width=0.22\textwidth]{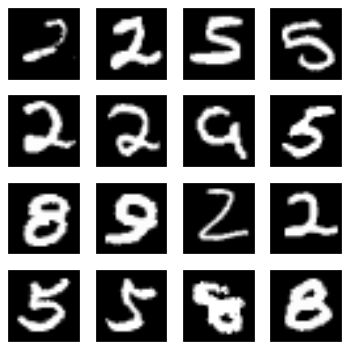}
    }\end{subfigure}
    \caption{
        FashionMNIST images with (a) the highest log-likelihood,
and    (b) the lowest log-likelihood.
  MNIST images with (c) the highest log-likelihood, and  
    (d) the lowest log-likelihood.
    }
    \label{fig:showimage_likelihood}
\end{figure}

\begin{figure}[ht]%
    \centering
       \begin{subfigure}[FashionMNIST: highest log-likelihood ratio.]{
       \includegraphics[width=0.25\textwidth]{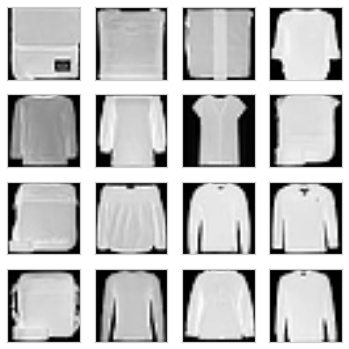}
        }\end{subfigure}
        \quad
       \begin{subfigure}[FashionMNIST: lowest log-likelihood ratio.]{
       \includegraphics[width=0.25\textwidth]{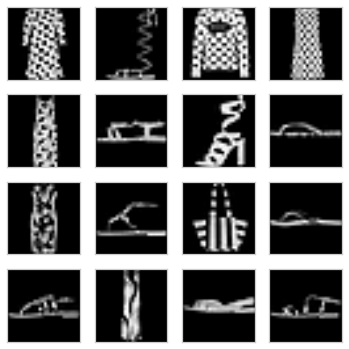}
        }\end{subfigure}
    \caption{
       FashionMNIST images with (a)  the highest and  
    (b) the lowest log-likelihood-ratios.
    }
    \label{fig:showimage_ratio}
\end{figure}

Images with the highest and lowest likelihood in Fashion-MNIST and MNIST dataset are shown in Figure~\ref{fig:showimage_likelihood}. 
Images with the highest likelihoods are mostly ``sandals'' in Fashion-MNIST dataset and ``1''s in MNIST dataset that have a large proportion of zeros. 
Images with the highest and lowest likelihood-ratio are shown in Figure~\ref{fig:showimage_ratio}.
Images with the highest likelihood-ratios are those with prototypical Fashion-MNIST icons such as ``shirts'' and ``bags'', highly contrastive with the background, while images with the lowest likelihood-ratios are those with rare patterns, such as dress with stripes or sandals with high ropes.

\begin{figure}[ht]%
    \centering
     \begin{subfigure}[]{
    \includegraphics[width=0.28\textwidth]{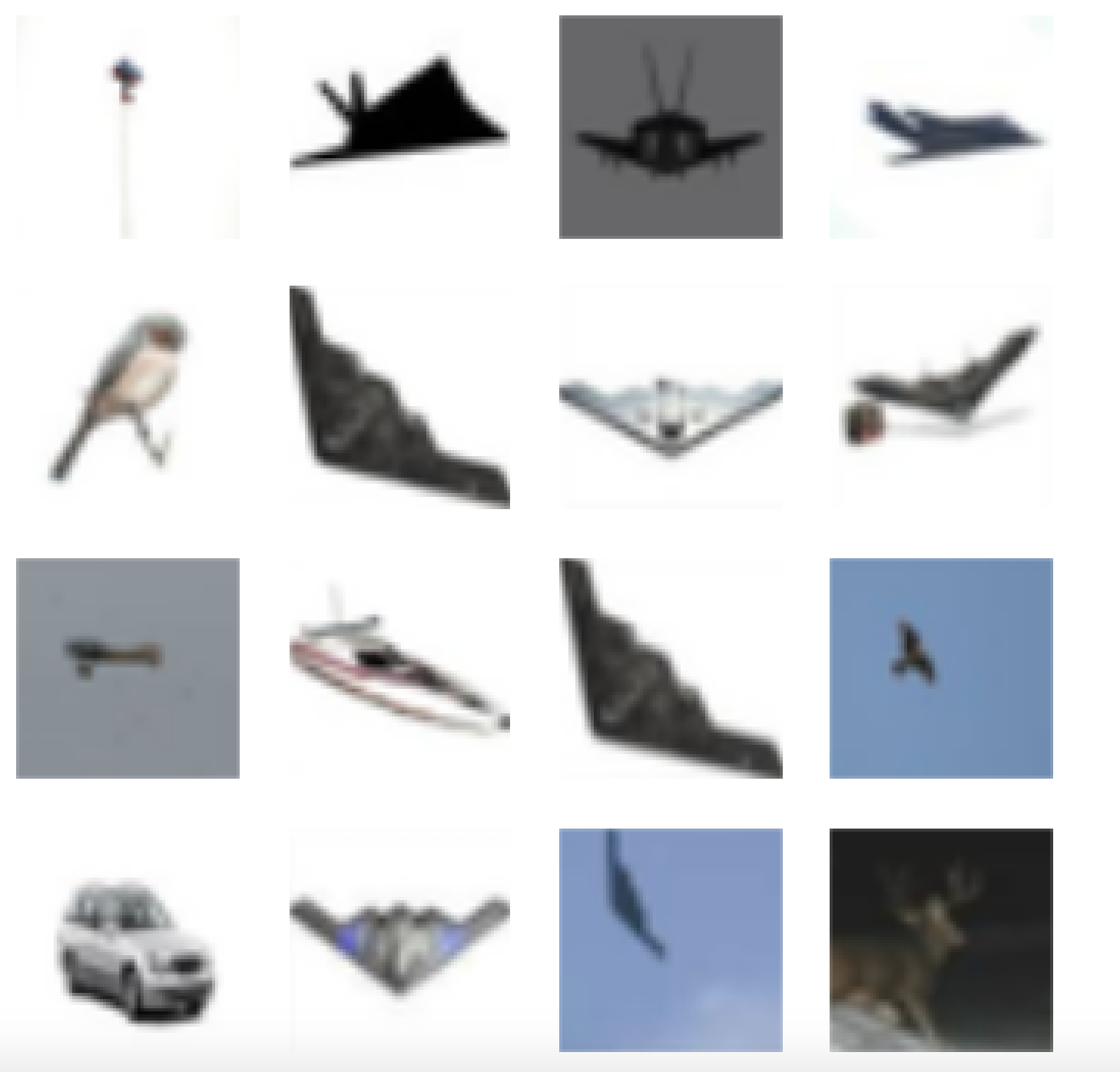}
    }\end{subfigure}
     \begin{subfigure}[]{
    \includegraphics[width=0.33\textwidth]{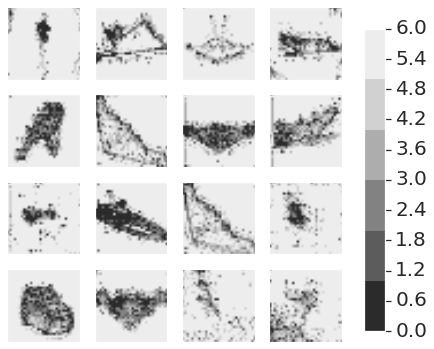}
    }\end{subfigure}
     \begin{subfigure}[]{
    \includegraphics[width=0.33\textwidth]{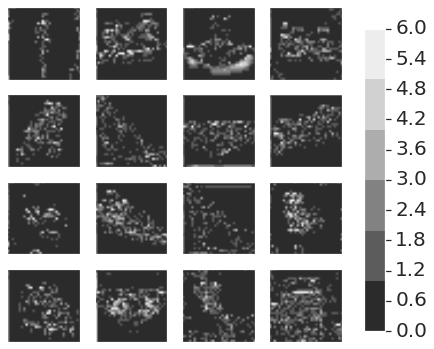}
    }\end{subfigure}
    \caption{%
    Examples of CIFAR-10 images (a), and their corresponding log-likelihood of each pixel in an image $\log p_{\vtheta}(x_d | \vx_{<d})$ (b), and the log likelihood-ratio of each pixel $\log p_{\vtheta}(x_d | \vx_{<d}) - \log p_{\vtheta_0}(x_d | \vx_{<d})$, are plotted as  $32\times32$ images.
    Lighter (darker) gray color indicates larger (smaller)  value. Note that the range of log-likelihood (negative value) is different from that of log likelihood-ratio (mostly positive value). For the ease of visualization, we unify the colorbar by adding a constant to the log-likelihood score. We picked the images that have the highest log-likelihood $p_{\vtheta}(\mathbf{x})$. }
    \label{fig:image_heat:cifar}
\end{figure}

Figure~\ref{fig:image_heat:cifar} shows qualitative results on CIFAR-10, displaying the per-pixel likelihood and likelihood-ratio as a heatmap. Similar to the trends in  Figure~\ref{fig:image_heat:mnist}, we observe that ``background'' pixels cause some CIFAR-10 images to be assigned high likelihood. 

\subsection{Supplementary tables}\label{sec:image_table}

The AUROC$\uparrow$, AUPRC$\uparrow$, and FPR80$\downarrow$ for CIFAR-10 (in-distribution) vs. SVHN (OOD) experiment is shown in Table \ref{tab:cifar}. Pure likelihood performs poorly for OOD detection with AUROC 0.095, and likelihood-ratio improves the performance significantly, achieving AUROC 0.931. Classifier-based ensemble methods perform well with AUROCs ranging from 0.937 to 0.946.
Note that our likelihood-ratio method is completely unsupervised whereas classifier methods require labels. Using likelihood-ratios on class-conditional generative models might further improve performance.

\begin{table}
\centering
\caption{
AUROC$\uparrow$, AUPRC$\uparrow$, FPR80$\downarrow$ for detecting OOD inputs using likelihood and likelihood-ratio method and other baselines on CIFAR-10 vs. SVHN datasets. Note that the generative-model based approaches (including our LLR method) are completely unsupervised, which puts them at a slight disadvantage, compared to classifier-based methods which have access to  labels. 
Numbers in front and inside of the brackets are mean and standard error respectively based on 10 independent runs with random initialization of network parameters and random shuffling of training inputs. For ensemble models, mean and standard error are estimated based on 10 bootstrap samples from 30 independent runs, which can be underestimations of the true standard errors. 
}
\vspace{-0.5em}
\begin{adjustbox}{max width=0.6\textwidth}
\begin{tabular}{llll}
& AUROC$\uparrow$ & AUPRC$\uparrow$ & FPR80$\downarrow$       \\ \hline
Likelihood & 0.095 (0.003) &	0.320 (0.001) & 1.000 (0.000) \\
Likelihood Ratio (ours, $\mu$) & 0.931 (0.032)	& 0.888 (0.049) & 0.062 (0.073) \\
Likelihood Ratio (ours, $\mu$, $\lambda$) & 0.930	(0.042)	& 0.881	(0.064)	& 0.066	(0.123) \\
$p(\hat{y}|\vx)$ & 0.910 (0.011) & 0.871 (0.019) & 0.094 (0.030) \\
Entropy of $p(\hat{y}|\vx)$ & 0.920 (0.013) & 0.890 (0.021) & 0.139 (0.020) \\
ODIN                    & 0.938 (0.091) & 0.926 (0.103) & 0.086 (0.179) \\
Mahalanobis distance    & 0.728 (0.108) & 0.711 (0.118) & 0.469 (0.178) \\
Ensemble, 5 classifiers & 0.937 (0.010) & 0.906 (0.017) & 0.037 (0.013) \\
Ensemble, 10 classifiers & 0.943 (0.004) & 0.915 (0.008) & 0.023 (0.002) \\
Ensemble, 20 classifiers & 0.946 (0.002) & 0.916 (0.004) & 0.020 (0.001) \\
Binary classifier       & 0.508 (0.027) & 0.505 (0.021) & 0.948 (0.107) \\
$p(y|x)$ with noise class & 0.923 (0.011) & 0.892 (0.016) & 0.064 (0.026) \\
$p(y|x)$ with calibration & 0.809 (0.043) & 0.735 (0.046) & 0.396 (0.101) \\ 
WAIC, 5 models & 0.146 (0.089)	& 0.343	(0.038) & 0.956 (0.062) \\ \hline
\end{tabular}
\end{adjustbox}
\label{tab:cifar}
\end{table}

\clearpage
\newpage
\section{Supplementary materials for the experiments on genomic sequences}

\subsection{Dataset design}\label{dataset}
We downloaded 11,672 bacteria genomes from National Center for Biotechnology Information (NCBI, \url{https://www.ncbi.nlm.nih.gov/genome/browse#!/prokaryotes/}) on September 2018.
For each genome we pooled its taxonomy information from the species level, to the genus, the family, the order, the class, and the phylum level.
High taxonomy levels such as the phylum level represents broad classification, while low taxonomy levels like the species and genus give a refined classification. 
To provide a precise classification, we use different genera as class labels, as has been done in previous studies \citep{brady2009phymm, ahlgren2016alignment}.
We filtered genomes that have missing genus information, or have ambiguous genus names.
A genus usually contains genomes from different species, subspecies, or strains. 
 \begin{figure}[ht]
\begin{center}
    \centering
\includegraphics[width=0.4\textwidth]{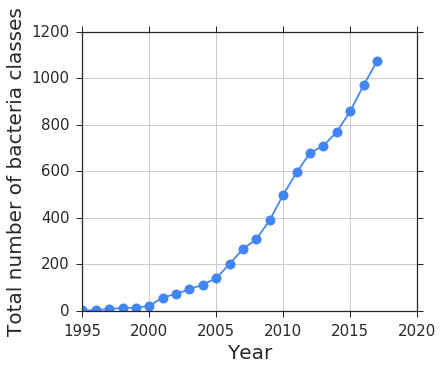} %
    \myvspace{-1em}
    \caption{Cumulative number of new bacteria classes discovered over the years (NCBI microbial genomes browser, September 2018.)}
    \label{fig:accumu}
\end{center}
\end{figure}
Different bacterial classes were discovered gradually over the years (Figure~\ref{fig:accumu}). 
Grouping classes by years is a natural way to mimic the in-distribution and OOD examples. 
Given a certain cutoff year, the classes discovered before the year cutoff can be regarded as in-distribution classes, and those after the year cutoff can be regarded as the OODs. 
In particular, we define the year that a class was first discovered as the earliest year when any of the genomes belonging to this class was discovered.
We choose two cutoff years, 2011 and 2016, to define the training dataset for in-distribution, validation datasets for in-distribution and OOD, and test datasets for in-distribution and OOD (Figure~\ref{fig:dataset}).
Genomes belonging to classes that were first discovered before 01/01/2011 are used for generating the training dataset for in-distribution.
Genomes belonging to new classes that were first discovered between 01/01/2011 and 01/01/2016 are used for generating the validation dataset for OOD.
Genomes belonging to the old classes but sequenced and released between 01/01/2011 and 01/01/2016 are used for generating the validation dataset for in-distribution. 
Similarly, genomes belonging to the new classes that were first discovered after 01/01/2016 are used for generating test dataset for OOD, while genomes belonging to the old classes that were sequenced and released after 01/01/2016 are used for generating the test dataset for in-distribution.
This setting avoids overlaps among genomes from training, validation, and test datasets. It is possible that different bacteria genomes may share similar gene regions, but those are rare for genomes from different genera and hence, we ignore this effect in our study.  
The bacteria class names are listed in Table \ref{tab:data_info}.

\begin{figure}[ht]%
    \centering
    \includegraphics[width=0.998\textwidth]{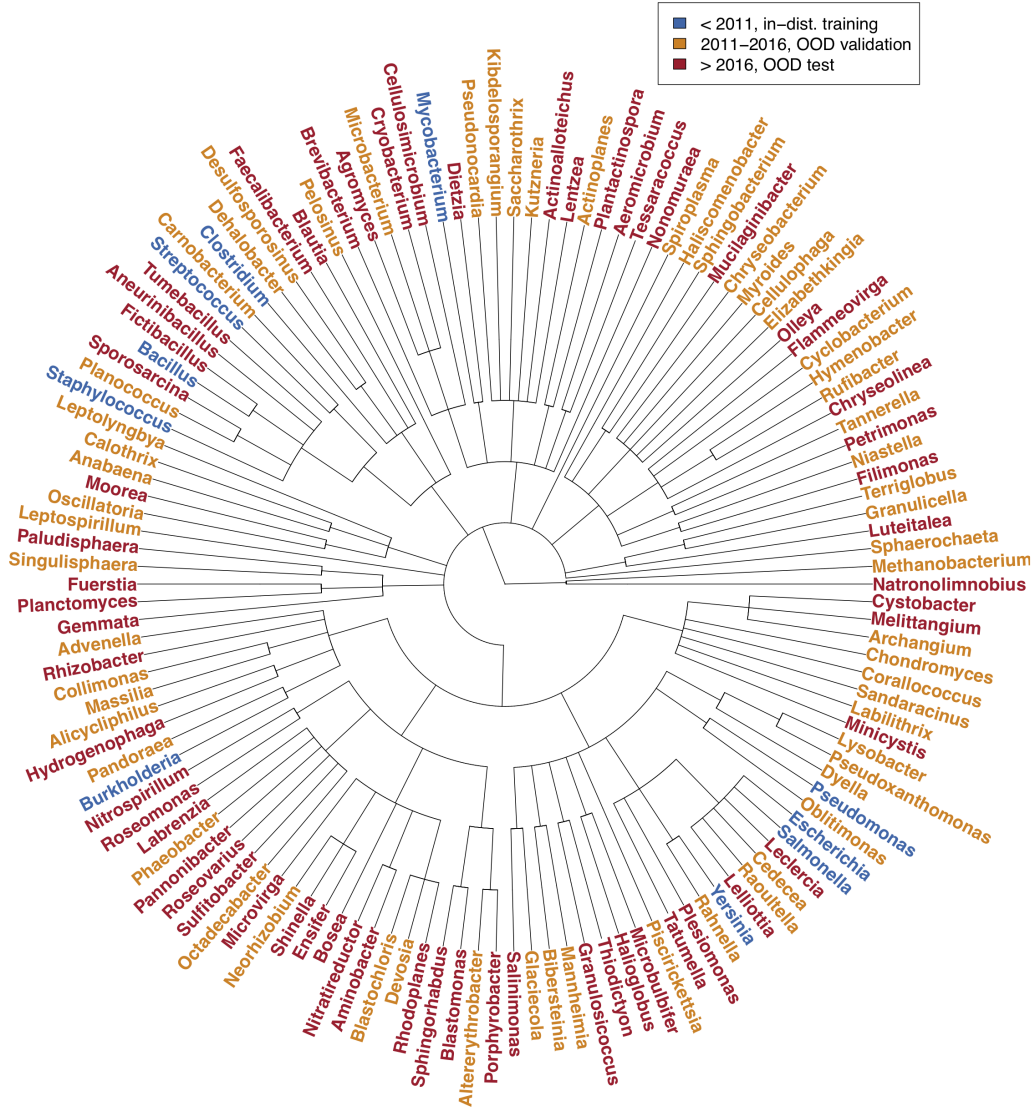}
    \caption{The phylogenetic tree of the 10 in-distribution, 60 OOD validation, and 60 OOD test bacteria classes. Note that the in-distribution and OOD classes are interlaced under the same taxonomy.}
    \label{fig:phylo}
\end{figure}

We designed a dataset containing 10 in-distribution classes, 60 OOD classes for validation, and 60 OOD classes for test.
The classes were chosen since they are the most common classes and have the largest sample sizes.
The in-distribution and OOD classes are interlaced under the same taxonomy (Figure \ref{fig:phylo}).
To mimic the real sequencing data, we fragmented genomes in each class into short sequences of length 250 base pairs, %
which is a common length that the current sequencing technology generates. 
Among all the short sequences, we randomly choose 100,000 sequences for each class for the training, validation, and test datasets.

\begin{table*}[htbp]
\center
\caption{The bacterial classes used in the genomic dataset for in- and out-of- distributions.}
\vspace{0.5em}
\begin{tabularx}{\linewidth}{ r X }
\hline
In-distribution training & Bacillus, Burkholderia, Clostridium, Escherichia, Mycobacterium, Pseudomonas, Salmonella, Staphylococcus, Streptococcus, Yersinia    \\ \hline
OOD validation           & Actinoplanes, Advenella, Alicycliphilus, Altererythrobacter, Anabaena, Archangium, Bibersteinia, Blastochloris, Calothrix, Carnobacterium, Cedecea, Cellulophaga, Chondromyces, Chryseobacterium, Collimonas, Corallococcus, Cyclobacterium, Dehalobacter, Desulfosporosinus, Devosia, Dyella, Elizabethkingia, Glaciecola, Granulicella, Haliscomenobacter, Hymenobacter, Kibdelosporangium, Kutzneria, Labilithrix, Leptolyngbya, Leptospirillum, Lysobacter, Mannheimia, Massilia, Methanobacterium, Microbacterium, Myroides, Neorhizobium, Niastella, Oblitimonas, Octadecabacter, Oscillatoria, Pandoraea, Pelosinus, Phaeobacter, Piscirickettsia, Planococcus, Pseudonocardia, Pseudoxanthomonas, Rahnella, Raoultella, Rufibacter, Saccharothrix, Sandaracinus, Singulisphaera, Sphaerochaeta, Sphingobacterium, Spiroplasma, Tannerella, Terriglobus \\ \hline
OOD testing              & Actinoalloteichus, Aeromicrobium, Agromyces, Aminobacter, Aneurinibacillus, Blastomonas, Blautia, Bosea, Brevibacterium, Cellulosimicrobium, Chryseolinea, Cryobacterium, Cystobacter, Dietzia, Ensifer, Faecalibacterium, Fictibacillus, Filimonas, Flammeovirga, Fuerstia, Gemmata, Granulosicoccus, Halioglobus, Hydrogenophaga, Labrenzia, Leclercia, Lelliottia, Lentzea, Luteitalea, Melittangium, Microbulbifer, Microvirga, Minicystis, Moorea, Mucilaginibacter, Natronolimnobius, Nitratireductor, Nitrospirillum, Nonomuraea, Olleya, Paludisphaera, Pannonibacter, Petrimonas, Planctomyces, Plantactinospora, Plesiomonas, Porphyrobacter, Rhizobacter, Rhodoplanes, Roseomonas, Roseovarius, Salinimonas, Shinella, Sphingorhabdus, Sporosarcina, Sulfitobacter, Tatumella, Tessaracoccus, Thiodictyon, Tumebacillus         \\       
\hline
\end{tabularx}
\label{tab:data_info}
\end{table*}

\subsection{Model details}\label{sec:gene_model}

For generative models of genomic sequences, we build a LSTM model \citep{lstm} to estimate the probability distribution of the next position given the history $p(x_d|\vx_{<d})$.
In particular, we feed the one-hot encoded DNA sequences into an LSTM layer, followed by a dense layer and a softmax function to predict the probability distribution over the 4 letters of $\{A, C, G, T\}$.
The model is trained using the in-distribution training data only. 
The size of the hidden layer in the LSTM was tuned via the in-distribution validation dataset and the final model uses 2,000 hidden units. The model is trained for 900,000 steps using learning rate of 0.0005, batch size of 100,  and Adam optimizer. The accuracy for predicting next character is 0.45 for in-distribution inputs.

We  train a background model by using the  perturbed in-distribution data 
and (optionally)  adding $L_2$ regularization to model weights. 
We search the optimal mutation rate $\mu$ from the range of $\mu=[0.01, 0.05, 0.1, 0.2]$, and 
evaluate the AUROC of 2,000 in-distribution and the same number of OOD inputs in the validation dataset.
Note that the set of OOD classes in the validation dataset is different from that in the test dataset. We tune hyperparameters without exposure to the final test OOD classes. 
The optimal $\mu$ is 0.2 with AUROC of 0.763 in validation data and 0.727 in the test dataset.
We also test if $L_2$ regularization helps for training the background model. 
Evaluating AUROC of OOD detection under different combinations of $\mu=[0.01, 0.05, 0.1, 0.2]$ and $\lambda=[0, 10^{-6}, 10^{-5}, 10^{-4}, 10^{-3}]$, we observe that AUROC is generally high for most of the combinations of the two hyperparameters except for some extreme cases (Table~\ref{tab:hyper_gene}a), when both $\mu$ and $\lambda$ are too high ($\mu \geq 10^{-3}$ and $\lambda \geq 0.2$) such that the model fails to learn informative patterns, or both are too small ($\mu \leq 10^{-6}$ and $\lambda \leq 0.05$) such that the background model is too similar to the in-distribution specific model. The optimal combination is $\mu=0.1$ and $\lambda=10^{-4}$, achieving AUROCs of 0.775 in validation dataset and 0.755 in the test dataset. %

\begin{table}[ht]
\caption{Hyperparameter tuning of mutation rate $\mu$ and $L_2$ coefficient $\lambda$ of the background model of our likelihood-raio method for genomic dataset. (a) Effects of the mutation rate $\mu$ and $L_2$ coefficient $\lambda$ on the AUROC$\uparrow$ for OOD detection of genomic sequences on the validation dataset containing 2,000 in-distribution and the same number of OOD inputs. When tuning only on mutation rate $\mu$, the optimal value is $\mu=0.2$. When tuning on both $\mu$ and $\lambda$, the optimal values are $\mu=0.1$ and $\lambda=10^{-4}$. (b) The same as (a) but tuning using simulated OOD inputs. The simulated OOD inputs are generated by permuting the in-distribution inputs at the mutation rate 10\%. The trend of the AUROC under different combintations of hyperparameters are similar with that using real OOD inputs. }
\begin{minipage}[b]{0.48\linewidth}
\centering
\begin{adjustbox}{max width=\textwidth}
\begin{tabular}{llllll}
& $\mu=0.01$     & 0.05     & 0.1      & 0.2 \\ \hline
$\lambda=0$        & 0.551 & 0.664 & 0.719 & \textbf{0.763} \\
$10^{-6}$ & 0.694 & 0.753 & 0.767 & 0.767 \\
$10^{-5}$  & 0.747 & 0.761 & 0.771 & 0.768 \\
$10^{-4}$  & 0.768 & 0.774 & \textbf{0.775} & 0.764 \\
$10^{-3}$  & 0.762 & 0.755 & 0.748 & 0.706 \\ \hline
\end{tabular}
\end{adjustbox}
\end{minipage}
\hfill
\begin{minipage}[b]{0.48\linewidth}
\centering
\begin{adjustbox}{max width=\textwidth}
\begin{tabular}{llllll}
& $\mu=0.01$       & 0.05  & 0.1   & 0.2       \\ \hline
$\lambda=0$      & 0.579 & 0.662 & 0.744 & \textbf{0.779} \\
$10^{-6}$ & 0.568 & 0.653 & 0.742 & 0.726 \\
$10^{-5}$ & 0.638 & 0.663 & 0.689 & 0.755 \\
$10^{-4}$ & 0.749 & 0.754 & \textbf{0.797} & 0.775 \\
$10^{-3}$ & 0.762 & 0.777 & 0.750 & 0.741 \\ \hline
\end{tabular}
\end{adjustbox}
\end{minipage}
\myvspace{1em}
\label{tab:hyper_gene}
\end{table}

We further study if the hyperparameters can be tuned without using the OOD inputs.
We use the perturbed in-distribution validation data as simulated OOD inputs. %
We choose the mutation rate as 10\%, because the average identity between bacteria is estimated 96.4\% at the genus level, and 90.1\% at the family level \citep{yarza2008all}.
Using the mutated in-distribution data to mimic OODs, we compute the likelihood-ratio for the in-distribution and the simulated OOD.
The optimal ranges of the hyperparameters under which high AUROC are similar with the previous choice based on the real OODs.
The optimal mutation rate when tuning without $L_2$ regularization ($\lambda=0$), and the optimal combination of the two hyperparameters, are the same as that tuned using real OOD input.

In order to compare with the classifier-based baselines, 
we build a classifier using convolutional neural networks (CNNs), which are commonly used in both image and genomic sequence classification problems~\citep{alipanahi2015predicting, zhou2015predicting, busia2018deep, ren2018identifying}. 
For genomic sequences, we feed one-hot encoded DNA sequence composed by $\{A, C, G, T\}$ into a convolutional layer, followed by a max-pooling layer and a dense layer. The output is then transformed to class probabilities using a softmax function. The number of filters, the filter size, and the number of neurons in the dense layer were tuned using the in-distribution validation dataset. This resulted in 1,000 convolutional filters of length 20 and 1,000 neurons in the dense layer. The accuracy of the classifier on the validation dataset is 0.8160.
Baseline methods 6-8 are based on perturbed in-distribution inputs, so the mutation rate is a hyperparameter for tuning. We use the same validation dataset as above, and tune the mutation rate ranging from $[0.0, 0.01, 0.05, 0.1, 0.2, 0.3, 0.4, 0.5]$.
For ODIN method, we tune the temperature and the input perturbation in the ranges of $[1, 5, 10, 100, 1000]$ and $[0, 0.0001, 0.001, 0.01, 0.1]$, respectively.

\subsection{Supplementary tables}

Table \ref{tab:corr} shows the minimum genetic distance between each of the OOD classes and in-distribution classes and its corresponding AUROC for OOD detection using Likelihood Ratio and classifier-based ensemble method with 20 models. We discovered that the AUROC for OOD detection is correlated with the genetic distance (Figure~\ref{fig:genomics}c).
The Pearson Correlation Coefficient are 0.570 for Likelihood Ratio method, and 0.277 for the ensemble method. 
The results confirm that in general a OOD class far away from the in-distribution is easier to be detected.

\begin{table}[ht]
\centering
\caption{Minimum genetic distance between each of the 60 OOD classes and in-distribution classes and their corresponding AUROCs for OOD detection.}
\vspace{0.5em}
\footnotesize{
\begin{tabular}{cccc}
OOD Class               & Min distance & AUROC$\uparrow$ (Likelihood Ratio) & AUROC$\uparrow$ (Ensemble 20) \\ \hline
Sulfitobacter      & 0.331     & 0.779    & 0.757           \\
Tumebacillus       & 0.323     & 0.784    & 0.814           \\
Blautia            & 0.320     & 0.708    & 0.828           \\
Roseovarius        & 0.319     & 0.747    & 0.756           \\
Moorea             & 0.293     & 0.904    & 0.819           \\
Natronolimnobius   & 0.273     & 0.857    & 0.585           \\
Fuerstia           & 0.266     & 0.887    & 0.841           \\
Chryseolinea       & 0.265     & 0.887    & 0.863           \\
Faecalibacterium   & 0.264     & 0.784    & 0.713           \\
Gemmata            & 0.260     & 0.864    & 0.615           \\
Aneurinibacillus   & 0.255     & 0.661    & 0.731           \\
Olleya             & 0.253     & 0.793    & 0.820           \\
Planctomyces       & 0.252     & 0.804    & 0.568           \\
Nitratireductor    & 0.250     & 0.759    & 0.694           \\
Filimonas          & 0.249     & 0.869    & 0.858           \\
Sphingorhabdus     & 0.249     & 0.852    & 0.845           \\
Mucilaginibacter   & 0.241     & 0.807    & 0.910           \\
Paludisphaera      & 0.240     & 0.846    & 0.632           \\
Petrimonas         & 0.240     & 0.898    & 0.878           \\
Flammeovirga       & 0.230     & 0.803    & 0.836           \\
Granulosicoccus    & 0.229     & 0.836    & 0.756           \\
Minicystis         & 0.225     & 0.783    & 0.493           \\
Labrenzia          & 0.224     & 0.720    & 0.724           \\
Microvirga         & 0.224     & 0.746    & 0.725           \\
Porphyrobacter     & 0.222     & 0.716    & 0.741           \\
Cellulosimicrobium & 0.217     & 0.760    & 0.601           \\
Agromyces          & 0.214     & 0.704    & 0.558           \\
Melittangium       & 0.209     & 0.824    & 0.692           \\
Cystobacter        & 0.208     & 0.745    & 0.632           \\
Blastomonas        & 0.207     & 0.782    & 0.777           \\
Pannonibacter      & 0.201     & 0.778    & 0.647           \\
Ensifer            & 0.201     & 0.750    & 0.758           \\
Nonomuraea         & 0.197     & 0.727    & 0.527           \\
Halioglobus        & 0.193     & 0.771    & 0.746           \\
Salinimonas        & 0.192     & 0.796    & 0.819           \\
Microbulbifer      & 0.190     & 0.790    & 0.791           \\
Roseomonas         & 0.189     & 0.706    & 0.618           \\
Plantactinospora   & 0.189     & 0.656    & 0.412           \\
Shinella           & 0.183     & 0.651    & 0.606           \\
Aeromicrobium      & 0.183     & 0.658    & 0.392           \\
Rhodoplanes        & 0.179     & 0.792    & 0.730           \\
Fictibacillus      & 0.179     & 0.742    & 0.738           \\
Bosea              & 0.178     & 0.693    & 0.722           \\
Rhizobacter        & 0.175     & 0.591    & 0.665           \\
Lentzea            & 0.175     & 0.733    & 0.609           \\
Brevibacterium     & 0.175     & 0.754    & 0.543           \\
Thiodictyon        & 0.173     & 0.773    & 0.698           \\
Plesiomonas        & 0.172     & 0.646    & 0.814           \\
Tessaracoccus      & 0.170     & 0.742    & 0.533           \\
Actinoalloteichus  & 0.165     & 0.706    & 0.425           \\
Sporosarcina       & 0.164     & 0.802    & 0.758           \\
Aminobacter        & 0.163     & 0.721    & 0.793           \\
Luteitalea         & 0.162     & 0.835    & 0.700           \\
Nitrospirillum     & 0.157     & 0.715    & 0.635           \\
Dietzia            & 0.147     & 0.796    & 0.306           \\
Tatumella          & 0.141     & 0.660    & 0.828           \\
Cryobacterium      & 0.138     & 0.736    & 0.554           \\
Hydrogenophaga     & 0.137     & 0.661    & 0.627           \\
Lelliottia         & 0.095     & 0.535    & 0.866           \\
Leclercia          & 0.094     & 0.512    & 0.807           \\  \hline
\end{tabular}
}
\label{tab:corr}
\end{table}

\end{document}